%% file: acl_latex.tex
\documentclass[11pt]{article}

\usepackage[final]{acl}

\usepackage{times}
\usepackage{latexsym}

\usepackage[T1]{fontenc}

\usepackage[utf8]{inputenc}

\usepackage{microtype}

\usepackage{inconsolata}

\usepackage{graphicx}

\usepackage{booktabs}
\usepackage{longtable}
\usepackage{tabularx}

\usepackage{subcaption}
\usepackage{placeins}

\title{Accelerated Drift of Scientific Prose in the LLM Era: \\
Evidence from an Interrupted Time-Series Analysis}

\author{Hanna Hubarava \\
  Department of Clinical Research\\
  Department of Digital Medicine\\
  University of Bern, Switzerland\\
  \texttt{hanna.hubarava@unibe.ch} \\\And
  Benjamin V. Ineichen \\
  Department of Clinical Research\\
  Department of Digital Medicine\\
  University of Bern, Switzerland\\
  \texttt{benjamin.ineichen@unibe.ch} \\}

\begin{document}
\maketitle
\begin{abstract}

Every scientific text is embedded in the writing norms of its time: it reflects earlier publications while also shaping the texts that follow. The growing use of LLM-based writing tools introduces a new factor in this evolving system, but the linguistic changes coinciding with their emergence remain insufficiently understood. We investigate whether the trajectories of scientific-preprint writing changed in rate or direction after conversational LLMs became publicly accessible, using the release of ChatGPT in November 2022 as the observational breakpoint. We compare pre- and post-2022 trajectories across vocabulary, punctuation, readability, and syntax in abstracts from arXiv’s artificial intelligence, statistics, and quantitative biology categories and from medRxiv. Using interrupted time-series regression, we find substantial post-2022 changes across multiple feature families and research domains, including increases in average word length, reduced formula-based readability, and declines in maximum dependency-tree depth and sentence-level depth variation. Individual features follow different dynamics: some lexical markers accelerate persistently, while others rise and then recede. 
Our findings show that isolated marker frequencies capture only part of the shift: changes in the rate, direction, and structure of linguistic trajectories provide complementary evidence of broader post-2022 change in scientific-preprint writing. 
More broadly, these findings motivate monitoring scientific writing as a multidimensional, evolving system rather than through a fixed set of LLM-associated markers.

\end{abstract}

\section{Introduction}\label{sec:intro}

Scientific writing constantly evolves as researchers adopt new terminology, rhetorical conventions, and ways of organizing information. The impact of such changes extends beyond style: scientific prose determines how efficiently readers can identify claims, recover relations between concepts, and compare findings across studies. Each abstract is shaped by prior publications and rhetorical habits of its community; once published, it enters the body of research from which future authors learn. Scientific writing is therefore self-referential: the texts it produces become evidence from which later researchers infer how science should be communicated in written form. 

Large Language Models (LLMs) have become increasingly integrated into this evolving linguistic landscape. They are trained, among other sources, on scientific text and increasingly used to draft, revise, and translate scientific manuscripts, and may then influence future writing norms and training data. Here, we ask whether the trajectories of linguistic features in scientific preprints changed around the onset of widespread public LLM availability, without attributing individual changes to LLM use.

Recent work has identified traces associated with LLM-mediated writing, including excess vocabulary \citep{kobak_delving_2025}, punctuation patterns \citep{freeburg2026fingerprintmarkdowntrainingshapes}, and other stylometric markers. Such markers are both salient and unstable: recognizable words may be avoided by authors, removed during editing, or change as models and prompting practices evolve.
The rise \citep{hern2024techscape} and subsequent fall \citep{geng_human-llm_2025} of words such as \textit{delve} illustrates why a frequency observed at a single point in time may misrepresent a developing linguistic shift (Fig.~\ref{fig:freq-examples}). Less conspicuous properties, such as readability, rhetorical stance and syntactic information packaging, may reveal more fundamental changes in the fabric of scientific writing overlooked by isolated marker counts \citep{shaib_detection_2024, zanotto_linguistic_2025}.


\input{figures/freq-examples}

We study abstracts from preprints spanning artificial intelligence, statistics, quantitative biology, and medicine. Their timestamps provide comparatively direct observations of when scientific texts entered public circulation, while their relatively stable communicative functions facilitate comparison across fields and years. Titles provide a secondary text-scope comparison of whether similar changes appear in an even more constrained form of scientific prose. 

Our analysis is limited to titles and abstracts rather than full texts, and to scientific preprints rather than peer-reviewed literature; our claims are restricted accordingly. To distinguish broad change from shifts concentrated in rapidly growing subjects, we compare both research domains and topic-level strata within the abstract corpora. Across these analyses, we track lexical markers, punctuation, readability, rhetorical stance, and dependency-based syntactic features.

 Our central contribution is to characterize linguistic change as a trajectory rather than a static pre–post difference. Using interrupted time-series models, we estimate whether feature trajectories changed after November 2022, which we treat as an intervention point marking the onset of widespread public access to conversational LLMs. We complement conspicuous lexical markers with measures of readability and dependency-based syntax. This allows us to examine changes in both surface form and the structural packaging of information. We compare these changes across research domains and topic-level strata to assess their breadth and heterogeneity, with titles providing a secondary comparison across text scopes.

\section{Related Research}\label{sec:related-research}

Research on the differences between human-written and AI-mediated text has often been framed as a detection problem. In this setting, linguistic and stylometric features are used as evidence for classifying texts as human- or machine-generated \citep{gao_comparing_2023, desaire_accurately_2023, zaitsu_stylometry_2025}. 
Our goal is instead to study population-level change: whether linguistic trajectories in scientific preprints shifted following widespread public access to LLMs.

Recent work has shown that LLM-mediated writing leaves detectable traces in scientific text, much of it lexical. \citet{liang_mapping_2024} estimate its prevalence from population-level word-frequency shifts, while \citet{kobak_delving_2025} identify ``excess vocabulary'' in biomedical abstracts. Related shifts appear in medical titles \citep{ibrahim_generative_2026}, cybersecurity papers \citep{vansteenhuyse_chatgpt_2026}, and academic writing by non-native English researchers \citep{lin_chatgpt_2025}. Punctuation may provide another signal \citep{freeburg2026fingerprintmarkdowntrainingshapes}. These studies inform our features, but their distributional markers do not reliably label individual documents.

LLM-associated change has been found to be uneven across scientific communities. \citet{liang_mapping_2024} and \citet{bao_examining_2025} report particularly strong changes in computer science, while \citet{bao_examining_2025} also report stronger reliance on LLMs for editing and proofreading among non-native English speakers, observing decreasing readability.
Unlike this prior work, we focus on changes in temporal trajectories and examine their heterogeneity across domains, topics, and text scopes.
Evidence that AI assistance may reduce stylistic and linguistic diversity \citep{agarwal_ai_2025,sourati_shrinking_2025} further motivates our analysis of structural variation, without implying a particular diffusion mechanism.

A limitation of marker-based analysis is that lexical traces may be unstable. \citet{geng_human-llm_2025} show that several words associated with ChatGPT, including \textit{delve}, rose sharply and then declined, interpreting this pattern as evidence of human-LLM coevolution. Such reversals may reflect human adaptation, changing prompting approaches, or shifts in model behavior across model families and post-training regimes \citep{freeburg2026fingerprintmarkdowntrainingshapes}. 
Structural properties provide complementary evidence: generated text exhibits recurring syntactic templates \citep{shaib_detection_2024}, reduced variation and increasing homogenization in model outputs \citep{zanotto_linguistic_2025}.

We therefore examine whether the \emph{velocity} and \emph{direction} of linguistic change altered following the public release of ChatGPT \citep{openai2022chatgpt}, rather than asking only whether particular markers became more frequent.

\section{Method}\label{sec:method}

\subsection{Study Design}

We conduct a diachronic comparative corpus study of scientific preprint abstracts and titles. We treat the public release of ChatGPT in November 2022 \citep{openai2022chatgpt} as an intervention point marking the onset of widespread access to conversational LLMs. Our primary analyses estimate changes in the trajectories of linguistic features in abstracts across four research corpora; matched analyses of titles provide a secondary text-scope comparison. We measure lexical markers, discourse markers, punctuation, readability, and dependency-based syntax. Analyses are performed separately within each corpus to prevent differences in corpus size from causing the largest domain to dominate the results. Topic-stratified analyses assess whether changes are confined to particular research topics. 

\subsection{Corpora and Temporal Coverage} \label{sec:method-corpora}
 For arXiv, we analyze preprints published from January 2015 through April 2026, with the data divided into three non-overlapping corpora: (i) artificial intelligence (\texttt{cs.CL}, \texttt{cs.AI}, and \texttt{cs.LG}, 403,247 records); (ii) quantitative biology (all \texttt{q-bio} categories, 38,424 records); and (iii) statistics (all \texttt{stat} categories, 115,255 records). The medRxiv corpus contains all available records from June 2019 through April 2026 (82,011 records). Across the four corpora, the analysis contains 638,937 records: 309,950 dated through 2022 and 328,987 dated from 2023 onward.

 For every record we collect the title, abstract, publication date, source identifier, and available subject metadata. We construct separate abstract and title datasets, producing eight corpus-by-scope datasets with matched record counts. We treat abstracts as the primary evidence base because they provide enough text for stable document-level estimates. Titles are analyzed as a distinct secondary scope, with sparse title features that cannot be estimated retained as missing rather than treated as null effects.


\subsection{Text Processing and Feature Extraction} \label{sec:method-processing-and-features}
We normalize text before feature extraction and process it with spaCy's English \texttt{en\_core\_web\_sm} pipeline \citep{honnibal_2020_spacy}. Features are first computed at document level and then aggregated by calendar month and year within each corpus and text scope. Lexical, discourse, phrase, and punctuation counts are normalized per 1,000 alphabetic words. Single-word markers were matched against lemmatized text; multiword expressions and punctuation were matched against the cleaned, non-lemmatized text. Appendix~\ref{app:data-collection-processing} provides more information on data collection and processing, while the complete list of operational definitions of the analyzed features is reported in Appendix Table~\ref{tab:features}.

\subsubsection{Lexical, Discourse and Punctuation Features} \label{sec:method-lexical-features}

The lexical inventory includes previously reported excess-vocabulary markers associated with LLM-mediated scientific writing \citep{kobak_delving_2025, liang_mapping_2024}, expressive verbs and adjectives, and selected multiword phrases. We analyze both individual items and lexical group totals. The inventory also includes sequential, causal, contrastive, emphasis, and summary discourse markers; separate certainty and hedging lexicons; and colons, semicolons, and em dashes. These distributions are population-level stylistic measures and are not used to classify individual documents as human- or machine-authored.

\subsubsection{Dependency-Based Syntactic Features} \label{sec:method-dependency-features}
For each document, we compute the distribution and entropy of dependency labels, mean dependency length and its across-sentence variation, maximum dependency-tree depth and variation in sentence-level tree depth, coordination count and its across-sentence variation, and the frequency of coordinated structures with exactly three conjuncts (list-of-three constructions). We additionally aggregate directed dependency-label bigrams and determiner contexts by year and period. These measures characterize changes in syntactic organization and information packaging.

\subsubsection{Readability Features}
We compute the following readability metrics using the \texttt{textstat} library: $\mu$ words per sentence, $\mu$ syllables per word, Flesch Reading Ease (FRE) \citep{flesch1948readability}, Flesch-Kincaid Grade Level (FKGL) \citep{Kincaid_Navy}, Dale-Chall score \citep{Chall_1995_ReadabilityR}, SMOG \citep{harry_smog_1969}, Automated Readability Index (ARI) \citep{ARI_smith}, and Gunning Fog Index \citep{gunning1952technique}. Because formula-based scores combine sentence- and word-level quantities, we additionally decompose changes in FKGL and ARI into their constituent components, which helps us identify which measured terms drive score changes.

\subsection{Topic-Level Heterogeneity}
We use topic analyses to assess heterogeneity. We perform them separately for each corpus, encode abstracts with \texttt{all-MiniLM-L6-v2} and fit BERTopic \citep{grootendorst_bertopic_2022}. 
We reduce embeddings to five dimensions with UMAP and cluster them with HDBSCAN. The minimum cluster size is 0.05\% of the corpus, with a lower bound of 10 documents; \texttt{min\_samples} is 10. Outliers are reassigned using BERTopic's embedding-based reduction procedure, and topics are reduced to at most ten per corpus. For interpretability, we label each topic by concatenating its top seven c-TF-IDF terms, avoiding manual relabeling. We then recompute feature trajectories and ITS models within topics.

\subsection{Interrupted Time-Series Analysis}
For each corpus, text scope, and feature,  we form a monthly series of document-level means.
\(y_t\) denotes the feature mean in month \(t\). We fit:

\[y_t = \beta_0 + \beta_1 t + \beta_2 I_t + \beta_3 T_t + \epsilon_t,\] 
where \(t\) is a month index, \(I_t\) equals one for months on or after November 2022 and zero otherwise, and \(T_t\) is zero before the intervention and counts months elapsed in the post-intervention period. Thus, \(\beta_0\) is the fitted baseline level, \(\beta_1\) is the pre-intervention monthly slope, \(\beta_2\) is the level shift, and \(\beta_3\) is the post-intervention change in slope. The post-intervention slope is therefore \(\beta_1+\beta_3\).

For comparability across features, we define the standardized slope-change estimate \(\Delta_{\sigma}\) as the annualized slope change divided by the pre-intervention standard deviation. Positive values indicate a shift toward a more positive trajectory, whereas negative values indicate a shift toward a more negative one. The magnitude \(|\Delta_{\sigma}|\) reflects the size of the trajectory change relative to pre-intervention variation.

Models are weighted by the number of papers in each month, and inference uses heteroskedasticity- and autocorrelation-consistent (HAC/Newey--West) standard errors with a maximum lag of six months. We adjust \(p\)-values using the Benjamini--Hochberg false-discovery-rate procedure within each feature family and corpus and report \(\Delta_{\sigma}\), 95\% confidence intervals, and adjusted \(q\)-values \citep{benjamini_controlling_1995}.

\subsection{Counterfactual Excess Analysis} \label{sec:method-counterfactual-excess}

Because the ITS model estimates a sustained change in slope, it may understate temporary or non-linear shifts. We complement it with counterfactual excess estimates for 2023--2024. For each feature, we fit a paper-count-weighted linear trend to pre-intervention months and extrapolate it through 2023--2024. We define mean counterfactual excess as the weighted mean difference between observed and projected monthly values during this period. We report standardized estimates, 95\% HAC confidence intervals, and within-family FDR-adjusted q-values.

\subsection{Cross-Domain and Text-Scope Comparisons} \label{sec:method-comparisons}

To assess whether linguistic changes are consistent across domains, we compare standardized slope-change estimates between corpora using Pearson correlations and directional agreement. We also compare titles with abstracts to determine whether the same changes occur across text scopes.

\section{Results}

\input{tables/main}


Table~\ref{table:main} shows the largest consistently significant effects across the four abstract corpora, while complete ITS estimates for all individual feature-corpus pairs are reported in Appendix Table~\ref{tab:full-feature-its}. 
Positive and negative \(\Delta_\sigma\) values indicate shifts toward more positive and more negative trajectories, respectively; they do not indicate whether a feature's absolute level increased or decreased. 
Titles serve as a secondary text-scope comparison (\S~\ref{sec:results-titles}).

The raw trajectories clarify why we distinguish sustained slope change from post-intervention excess (Fig.~\ref{fig:freq-examples}). \textit{Across} and syllables per word rise persistently after 2022, whereas \textit{insight} peaks and then recedes; maximum dependency-tree depth shifts from a flat or rising trajectory to a decline. ITS quantifies the sustained change relative to the prior slope, while counterfactual excess (\S~\ref{sec:results-counterfactual-excess}) captures the cumulative departure over a specified post-intervention window, including temporary trajectory departures (see Appendix Table~\ref{tab:full-feature-excess-2023-2024}).

\subsection{Lexical and Punctuation Markers} \label{sec:results-lexical-and-punctuation}

Aggregate marker-word frequency had a positive slope change in every corpus, from $2.78$ in medRxiv to 3.94 in AI (all \(q<.001\)). The most consistent individual marker was \textit{across}, with standardized changes of 6.10 in AI, 3.88 in quantitative biology and statistics, and 3.10 in medRxiv (all \(q<.001\)); \textit{while} also showed positive slope changes in every corpus. Expressive verbs show similar heterogeneity. \textit{Align} shows a positive slope change in all four corpora, with standardized changes of 3.14 in AI, 1.51 in quantitative biology, 1.92 in statistics, and 2.71 in medRxiv (all \(q<.001\)). \textit{Underscore} produces the largest verb effect in medRxiv (4.53, \(q<.001\)) but is not significant in any arXiv corpus. \textit{Delve} rises sharply but then recedes, giving negative but nonsignificant slope-change estimates. Adjective and multiword-phrase categories generally rise descriptively, but many peak and then reverse.

Punctuation changes were less uniform (Fig.~\ref{fig:main-punctuation-forest}): em-dash use shows a positive slope change in all four corpora, with the largest standardized effect in medRxiv (2.84, \(q<.001\)); effects in the three arXiv domains ranged from 1.45 to 2.35 (all \(q<.001\)). Colon shows a significant acceleration across all arXiv corpora, while semicolon trajectories varied more.
\input{figures/main-punctuation-forest}

\subsection{Change in Readability}

All four corpora show increased formula-based readability complexity after 2022 (Fig.~\ref{fig:main-readability-forest}). Standardized FKGL slope changes were $3.44$ pre-intervention SDs/year in AI, $2.79$ in quantitative biology, $1.85$ in statistics, and $0.66$ in medRxiv (all \(q<.001\)). In raw terms, these correspond to additional slope changes of $0.54$, $0.55$, $0.36$, and $0.17$ grade units/year. ARI changes were $2.93$, $3.00$, $1.97$, and $0.70$ SDs/year (all \(q<.01\)). FRE changes were $-3.50$, $-3.41$, $-2.56$, and $-0.87$ SDs/year (all \(q<.001\)). Effects were larger in arXiv, but \textit{directionally consistent} across all four corpora.

\input{figures/main-readability-forest}

\input{figures/main-readability-decompose}

As seen in Fig.~\ref{fig:main-readability-decompose}, the observed post-2022 drop in automatic readability can be linked mainly to longer words. Syllables/word account for the largest component of the FKGL slope change: $81.7\%$ in AI, $93.9\%$ in quantitative biology, $80.2\%$ in statistics, and $53.9\%$ in medRxiv. Sentence length contributes only $0.9$--$13.2\%$ in the arXiv corpora and changes in the opposite direction in medRxiv. For ARI, its implied characters/word component accounts for $70.1$--$99.2$\% of the absolute component change.

\subsection{Syntactic Structure and Information Packaging}
Maximum dependency-tree depth has a negative slope change in every corpus: $-2.13$ in AI, $-1.31$ in quantitative biology, $-1.69$ in statistics, and $-1.88$ in medRxiv (all \(q<.001\)). Variation in dependency-tree depth likewise declined, as did dependency-role entropy, although the entropy effects were smaller. 
Other syntactic measures were less consistent: coordination slopes became more positive in arXiv but more negative in medRxiv, while lists of three had their largest positive slope changes in arXiv's AI and statistics.
\input{figures/main-syntax-forest}

Dependency context provides additional evidence of changing information packaging (Fig.~\ref{fig:main-dependency-trends}). Post-intervention, determiners attached to prepositional-object heads fell from $40.6$ to $24.5$ per $1,000$ words in AI, $42.9$ to $28.6$ in quantitative biology, $46.5$ to $36.3$ in statistics, and $33.4$ to $22.1$ in medRxiv. Over the same period, the ratio of adjectival and compound prenominal modifiers to prepositional-modifier roles increased. These aggregate patterns are consistent with less prepositional elaboration and relatively more compact prenominal noun-phrase modification, but they do not imply that one construction replaced another or caused readability changes.

\input{figures/main-dependency-trends}

\subsection{Rhetorical and Discourse Markers}

Rhetorical changes were domain dependent. Hedging declined in all three arXiv corpora but increased in medRxiv. Certainty accelerated significantly in statistics and medRxiv, but not in AI or quantitative biology after correction. Emphasis markers changed most strongly in medRxiv ($2.47$, \(q<.001\)). The results therefore do not support a universal shift toward greater assertiveness. Importantly, these measures are lexicon based and do not capture rhetorical function in context.



\subsection{Topic-Level Heterogeneity}

Fig.~\ref{fig:main-topics-heatmap} shows that the aggregate results are not confined to changes in the prevalence of a single topic. The largest effects occur in arXiv AI topics which appear to be associated with deep learning and LLMs (see topic labels in Appendix~\ref{app:topic-trajectories}). However, similar directional changes appear in non-LLM topics and in the quantitative-biology, statistics, and medical corpora. MedRxiv displays a broad but generally weaker distribution of effects across topics. Because topics were estimated independently within each corpus, topic identifiers are not matched across domains. The analysis therefore shows that aggregate trajectories are not confined to a single within-corpus topic; it does not quantify how much of each corpus-level change is caused by changing topic prevalence.

\input{figures/main-topics}

\subsection{Cross-Corpus Similarity}
Despite differences in individual effects, the overall profiles of post-intervention slope changes are similar across corpora (Appendix~\ref{app:cross-corpus-similarity}; Fig.~\ref{fig:corpus-correlation-heatmap}). Pairwise correlations are high among the arXiv corpora ($r=.93$--$.95$), while medRxiv shows more moderate similarity with arXiv ($r=.69$--$.74$). This indicates cross-corpus similarity in the pattern of trajectory changes, without implying equal effect magnitudes or significance across domains.


\subsection{Titles as a Secondary Text-Scope Comparison} \label{sec:results-titles}

Title changes only partially aligned with abstract changes (Appendix~\ref{app:title-abstract-comparison}; Fig.~\ref{fig:main-titles}). Across features estimable in both scopes, agreement in the direction of slope change ranged from $51\%$ in q-bio to $74\%$ in AI. \textit{Across}, colon use, and maximum tree depth changed in the same direction across all corpora, whereas em-dash use showed opposite slope changes, shifting more positively in abstracts and more negatively in titles. Readability patterns also differed by scope: FRE showed negative slope changes in both titles and abstracts across the arXiv corpora, but a negative slope change in medRxiv abstracts and a positive one in its titles. Syntactic agreement was strongest in medRxiv, where the slopes of all syntactic features changed in the same direction. Overall, title effects were generally smaller and not consistently aligned with abstract effects.

\subsection{Counterfactual Excess Analysis} \label{sec:results-counterfactual-excess}
Counterfactual excess complements ITS by capturing temporary or nonlinear departures from the projected pre-intervention trajectory. Excluding group aggregates, significant excess occurred in 174 out of 304 feature–corpus estimates. Lexical features frequently departed from their projections without sustained change in slope, whereas some syntactic changes were detected primarily by ITS. Post-intervention change comprised both early deviations from prior trends and persistent changes in linguistic trajectories (complete feature-level estimates: Appendix Table~\ref{tab:full-feature-excess-2023-2024}).

\section{Discussion}

Our results show that the shift in scientific-preprint abstracts following widespread public LLM availability extends beyond conspicuous vocabulary and punctuation. Across four corpora, we observe changes in the trajectories of lexical markers, formula-based readability, and syntactic structure. These changes take different temporal forms: some features accelerate persistently, others show temporary elevations or reversals, and their magnitude varies substantially across domains and topics. Together, the results suggest that the linguistic changes coinciding with widespread LLM availability are broader than the rise of a small set of recognizable markers.

\paragraph{From marker words to broader stylistic change.} 

Prior research on shifts in marker-word frequencies provides the foundation for our study. Our trajectory-based approach shows that these markers follow different temporal patterns: \textit{across} and \textit{align}, for example, exhibit sustained positive slope changes, whereas \textit{delve}, \textit{insight}, and other salient items rise and subsequently recede. Examining the timing and persistence of such changes complements comparisons of average frequencies. 
By capturing both temporary departures from earlier trends and sustained trajectory changes, our analyses extend marker-based findings toward a broader account of post-2022 stylistic change.

\paragraph{Readability and lexical complexity.}

Readability provides one of the clearest cross-domain patterns. All four abstract corpora show post-2022 trajectory changes toward greater formula-based complexity; the effects are stronger in arXiv than in medRxiv. Decomposing the readability formulas shows that the changes are driven primarily by increasing word length rather than sentence length: syllables per word account for most of the FKGL slope change, and the corresponding word-length component similarly dominates ARI. The result is therefore more specific than a general increase in textual complexity. 

This distinction matters because abstracts are often the first (and sometimes only) part of a paper that readers encounter. A shift toward more lexically complex abstracts could have consequences for the accessibility of scientific findings, including beyond specialist audiences. However, readability formulas are only proxies for processing difficulty. Our results establish a change in measurable textual properties, not that readers necessarily find post-2022 abstracts harder to read. Reader studies are needed to determine whether these changes affect reading time or comprehension.

\paragraph{Changes in syntactic information packaging.}
The syntactic results do not indicate a simple shift toward either greater or lesser syntactic complexity. After 2022, the trajectories of maximum tree depth and tree-depth variation shift downward across all corpora, as does that of dependency-role entropy, while slope changes in coordination and dependency length are less consistent. Complementary descriptive trajectories show that determiner contexts involving prepositional objects become less frequent, while adjectival and compound prenominal modification increases relative to prepositional modification. These patterns suggest a shift towards reduced maximum dependency-tree depth and a greater relative role for compact noun-phrase modification.

The combination of longer words with reduced variation in some syntactic measures is notable because lexical elaboration need not imply greater structural diversity. Prior work has reported recurring syntactic templates and reduced linguistic variation in generated text \citep{shaib_detection_2024, zanotto_linguistic_2025}. Our population-level results are compatible with such regularization, but do not directly demonstrate template reuse or homogenization.

\paragraph{Uneven changes across domains, topics, and text scopes.}
The changes are not evenly distributed across science. Consistent with previous work \citep{liang_mapping_2024, bao_examining_2025}, some of the largest effects occur in arXiv AI topics associated with deep learning and language models. This concentration could reflect greater exposure to or earlier adoption of LLM-based writing tools, but our data cannot establish that mechanism. Importantly, the observed changes are not confined to LLM-related research. Comparable directional shifts appear in other AI topics and across quantitative biology, statistics, and medicine. Similarly, topic-level stratification shows that the corpus-level results are not driven exclusively by a single rapidly growing topic within each corpus. 
The heterogeneity itself is informative: the results point to changes that differ in strength and sometimes direction across scientific communities.

Titles provide a useful secondary comparison. Their trajectories only partially align with those of abstracts, and some features behave differently across the two scopes. This divergence is unsurprising given the different communicative constraints of titles and abstracts, and highlights the need to treat them separately. More importantly, it cautions against generalizing the abstract-level results to scientific prose as a whole.

\paragraph{Interpretation and implications.}
November 2022 provides a meaningful temporal breakpoint because it marks the onset of widespread public access to conversational LLMs, but it is not a randomized intervention; temporal alignment does not establish that LLM use caused the observed changes. 
At the same time, temporal placebo checks showed that earlier intervention dates often captured pre-existing drift, but the November 2022 breakpoint produced the largest absolute slope change in 19 of 20 representative feature–corpus comparisons (Appendix~\ref{app:placebo-checks}; Table~\ref{tab:placebo}).
Direct drafting and editing with LLMs are plausible contributors, but so are machine translation, changing editorial practices, or broader stylistic trends and fashions. These mechanisms may also interact: stylistic patterns introduced or amplified through LLM-mediated writing can enter the scientific literature and subsequently become part of the linguistic environment encountered by both authors and future models. For the same reason, the features examined here should not be treated as fingerprints of AI-assisted writing: their value is at the population level. 

The broader question is therefore not simply whether LLMs leave detectable traces, but rather how conventions in preprint abstracts evolve in an environment where researchers increasingly write with systems trained on earlier scientific writing. Our results provide evidence of a change, while leaving its mechanisms and consequences open.

\section{Conclusions}
We investigated whether the trajectories of scientific-preprint abstracts changed following widespread public access to conversational LLMs. Across four abstract corpora, post-2022 changes extend beyond conspicuous lexical markers to formula-based readability and syntactic information packaging. The most consistent patterns include longer words and reduced formula-based readability, lower maximum dependency-tree depth, and reduced sentence-level depth variation. These changes are strongest in parts of arXiv associated with AI and language models, but are not confined to these topics. With varying magnitude, they are observed across scientific domains.

The observed changes do not follow a single temporal pattern. While some lexical markers show sustained acceleration, others rise sharply and subsequently recede. Scientific linguistic change in the LLM era therefore cannot be reduced to a stable inventory of marker words: changes in the rate, direction, and structure of linguistic trajectories provide complementary evidence of a broader shift in scientific-preprint writing.

Together, these results shift the question from whether and how often LLM-associated marker words appear in scientific text to how the trajectories of scientific writing are changing. Determining the precise mechanisms behind these changes and whether they affect the accessibility, diversity, or communication of science will require linking population-level linguistic trends to writing practices and reader outcomes.

\section*{Limitations}\label{sec:limitations}

\paragraph{Intervention point and causal attribution.}
November 2022 is an observational breakpoint marking the onset of widespread public access to conversational LLMs, not a randomized intervention or a point at which their use became immediately widespread. Adoption likely increased gradually and unevenly across regions, disciplines, and research communities. The interrupted time-series approach identifies changes in linguistic trajectories around this point but cannot attribute them to LLM use. Changes in scientific communities, editorial practices, machine translation, or other writing technologies may all contribute to the observed patterns. Far from claiming that the observed changes are attributable to generative AI alone, we interpret the results as post-2022 linguistic changes coinciding with increasing LLM availability rather than estimates of a causal LLM effect.

\paragraph{Corpus scope.}
Our primary analyses are restricted to preprint abstracts. Our results therefore capture the most visible and compressed part of scientific writing and might not generalize to full-text preprints. Titles provide a secondary text-scope comparison, while full texts are not analyzed. Preprints have not necessarily undergone the editorial and peer-review processes that may affect published manuscripts. Our findings therefore should not be generalized to full papers or scientific writing in general. 

\paragraph{Feature validity.}
Some feature families are imperfect proxies for the linguistic properties they are supposed to characterize. Formula-based readability scores do not directly measure human comprehension, and lexical inventories for certainty, hedging, and other rhetorical categories measure frequencies of preselected expressions rather than their contextual function. Similarly, dependency-based measures do not directly establish syntactic templating or overall linguistic homogenization. Future work could combine these population-level measures with contextual annotation and reader studies.

\paragraph{Dependency parser.} Dependency-based measures rely on spaCy's \texttt{en\_core\_web\_sm} parser and were not manually validated on our corpora. While uniform processing supports comparisons, parsing errors may affect dependency labels, tree depth, and edge-bigram patterns. These results should therefore be interpreted only as parser-based proxies for syntactic change.

\paragraph{Topic stratification.} 
Topic-level stratification shows that the aggregate changes are not confined to a single rapidly growing topic within each corpus, but it does not fully separate within-topic linguistic change from changes in corpus composition. BERTopic partitions are sensitive to model choices, including embedding, dimensionality reduction, clustering, and topic reduction parameters, while c-TF-IDF labels provide only approximate cluster descriptions. Topic-level results should therefore be interpreted as evidence of heterogeneity rather than as a causal decomposition of corpus-level change.

\section*{Ethical considerations}\label{sec:ethics}
Our main goal is to capture and quantify the trajectories of change in scientific writing. We do not attempt to identify individual papers, authors, institutions, or countries as using generative AI for academic writing. The features studied here should not be interpreted as evidence of LLM use at the document level: they may also reflect human idiosyncrasies, light editing, translation support, disciplinary style, non-native English writing, or ordinary linguistic shifts. 

We caution against using our findings for individual-level AI detection or screening. Because public discussion of ``LLM markers'' may stigmatize lexical choices or disproportionately target non-native English researchers \citep{liang_gpt_2023} who are known to rely more heavily on LLMs for editing \citep{bao_examining_2025}, we report results at corpus and topic levels only.

\paragraph{Artifact Use.}
We use publicly available preprint metadata and abstracts from arXiv and medRxiv in accordance with their access policies. The analysis pipeline relies on open-source Python libraries, including \texttt{spaCy}, \texttt{BERTopic}, and \texttt{textstat}.

\paragraph{AI Assistance Disclosure.}
We used AI assistants for coding support, debugging, manuscript revisions, including translation and proofreading, as well as formatting tables and figures in LaTeX. All analyses, interpretations, and final manuscript decisions were reviewed and edited by the authors.


\section*{Lay Summary} \label{sec:lay-summary}
Large language models such as ChatGPT are increasingly used to help researchers write and edit scientific papers. Previous studies looked for signs of this use, often by asking whether particular words or punctuation grew more frequent after LLMs became widely available. We ask a different question: did the \emph{rate and direction} of change in scientific writing itself shift? In other words, rather than comparing how common a linguistic feature was before and after 2022, we ask whether it began increasing or decreasing faster, more slowly, or in a different direction than before.

We analyze more than 600,000 scientific-preprint abstracts from computer science, statistics, quantitative biology, and medicine. We track the trajectories of vocabulary, punctuation, readability, and the way sentences are built before and after 2022 (ChatGPT was released in November 2022). We find that changes in these trajectories extend beyond a few salient words or phrases. For example, across research fields, abstracts increasingly use longer words and score as less readable according to standard readability formulas. Some changes persist, while others temporarily depart from their earlier trajectories and then recede or reverse.

Although these patterns coincide with the onset of widespread access to LLMs, our study cannot show that LLM use caused them. Other changes in scientific communities, editing practices, translation tools, and writing conventions may also contribute. Our results suggest that tracking how the rate and direction of linguistic change evolve over time can complement studies of individual words and provide a broader picture of how scientific writing is changing.

\bibliography{custom}


\appendix









\section{Software and Computational Resources}\label{sec:appendix-compute-resources}

\paragraph{Code.} The analysis code, feature definitions, and environment lockfile are available at \url{https://github.com/shtosti/accelerated-drift}. Version 1.0.0 is permanently archived on Zenodo \citep{hubarava2026accelerateddrift}.

\paragraph{Software.} 
Analyses were implemented in Python 3.11. Topic embeddings used \texttt{all-MiniLM-L6-v2}; clustering used BERTopic with five-dimensional UMAP, embedding-based outlier reassignment, and reduction to at most ten topics per corpus. Topic labels concatenate the seven highest c-TF-IDF terms. Package versions are recorded in the released lockfile.

\paragraph{Hardware.} 
Calculations were performed on \href{https://www.id.unibe.ch/hpc}{UBELIX}, the HPC cluster at the University of Bern. We acknowledge the University of Bern for providing these computational resources. 

Full analysis jobs were submitted through SLURM using 16 CPU cores, 80 GB of system memory, and a single NVIDIA A100 or H100 GPU, depending on availability. The GPU was used to compute topic-model embeddings; feature extraction, clustering, statistical analysis, and visualization otherwise ran on CPU.

\section{Data Collection and Preprocessing} \label{app:data-collection-processing}
\subsection{Corpus construction}
We collected preprint metadata from arXiv and medRxiv through 30 April 2026. The arXiv corpora were defined by submission categories: artificial intelligence (\texttt{cs.CL, cs.AI, cs.LG}), quantitative biology (all \texttt{q-bio} categories), and statistics (all \texttt{stat} categories). A record assigned to more than one category within a corpus was retained once using its source identifier. The medRxiv corpus included all records available in the collection interval. The primary analysis used abstracts; titles were processed separately as a secondary text scope. We did not analyze full texts.

The four corpora contain 638,937 records: 403,247 arXiv AI, 38,424 arXiv quantitative biology, 115,255 arXiv statistics, and 82,011 medRxiv records. Of these, 309,950 were dated through 2022 and 328,987 from 2023 onward. arXiv coverage begins in January 2015 and medRxiv coverage in June 2019.

\subsection{Text Normalization}
Text was normalized before feature extraction and processed with \texttt{spaCy}'s English \texttt{en\_core\_web\_sm} pipeline.  Features were computed at document level before aggregation; the statistical models used monthly document-level means, with months weighted by paper counts. Title and abstract datasets retain matched records but are analyzed independently. When a feature cannot be estimated for a short title, it remains missing and is not replaced by zero. Table~\ref{tab:features} reports operational definitions of all feature groups and their respective units of analysis.

\input{tables/features}

\section{Features and Implementation} \label{app:features-and-implementation}
The analysis contains 76 individual features. Group-aggregate variables are excluded from the reported tables, significance counts, and within-family multiple-testing correction. Family assignments and operational definitions are reported in Table~\ref{tab:features}. Frequency features are measured per 1,000 alphabetic words; ratios, entropy measures, dependency measures, and readability scores retain their native scales before standardization.

\subsection{Lexical, Discourse, and Punctuation Measures}

The lexical inventory includes 11 marker words, 8 expressive verbs, 5 adjectives, and 6 multiword phrases. Discourse measures include five sequential, five causal, five contrast, five emphasis, and five summary markers. Group totals are not included in inference. Certainty and hedging are lexicon-based ratios. Punctuation measures count em dashes, colons, and semicolons.

\subsection{Readability Measures}

We computed mean words per sentence, mean syllables per word, Flesch Reading Ease, Flesch-Kincaid Grade Level, Dale-Chall, SMOG, Automated Readability Index, and Gunning Fog using the \texttt{textstat} library in Python. Because several formulas combine sentence and word length, we decomposed the observed FKGL and ARI slope changes into their component terms (Fig.~\ref{fig:main-readability-decompose}).

\subsection{Dependency-Based Syntax}
\label{appendix-dependency-syntax}

For each sentence, dependency-tree depth was defined recursively as one plus the maximum depth of its children. Terminal tokens were assigned depth one. Document-level maximum tree depth was the largest sentence depth in the document. Variation in sentence depth was the sample standard deviation across sentence depths.

Dependency length was the absolute difference between a token's and its syntactic head's spaCy token indices, averaged over all non-root dependency arcs in the document. Dependency-label entropy was Shannon entropy over the document's spaCy dependency-label distribution after excluding tokens labelled \texttt{punct}; we retained the \texttt{ROOT} label.

Coordination count was the number of tokens labelled \texttt{cc}. Coordination variation was the sample standard deviation of sentence-level \texttt{cc} counts. A list of three was counted when a token \textit{and} or \textit{or} labelled \texttt{cc} headed a coordination structure containing exactly three conjuncts.

\subsubsection{Dependency Roles and Labels}

Dependency labels describe the grammatical relation between a token and its syntactic head in the spaCy parse. For the role-level analysis, we calculated each label's yearly proportion among non-punctuation dependency tokens. Roles were ranked by their mean absolute post-2022 slope change across the four abstract corpora, and the twelve largest were retained.

\begin{center}
\small
\resizebox{0.7\linewidth}{!}{%
\begin{tabular}{@{}ll@{}}
\toprule
Label & Meaning \\
\midrule
\texttt{ROOT}     & sentence root \\
\texttt{det}      & determiner \\
\texttt{prep}     & prepositional modifier \\
\texttt{auxpass}  & passive auxiliary \\
\texttt{conj}     & conjunct \\
\texttt{cc}       & coordinating conjunction \\
\texttt{dobj}     & direct object \\
\texttt{pobj}     & object of preposition \\
\texttt{nsubj}    & nominal subject \\
\texttt{aux}      & auxiliary verb \\
\texttt{advcl}    & adverbial clause modifier \\
\texttt{compound} & compound modifier \\
\texttt{amod}     & adjectival modifier \\
\texttt{npadvmod} & noun-phrase adverbial modifier \\
\texttt{nummod}   & numeric modifier \\
\texttt{appos}    & appositional modifier \\
\bottomrule
\end{tabular}%
}
\end{center}

\subsubsection{Noun-Phrase Context Measures}

To contextualize changes in individual dependency roles, we examined two aggregate measures of noun-phrase modification. First, we counted \texttt{det} tokens whose syntactic head was labelled \texttt{pobj} and reported their frequency per 1,000 words. Second, we calculated the ratio of prenominal-modifier roles (\texttt{amod} and \texttt{compound}) to prepositional-modifier roles (\texttt{prep} and \texttt{pobj}). These measures provide context for the results reported in Fig.~\ref{fig:main-dependency-trends}, but do not identify matched substitutions between constructions.

Figure~\ref{fig:main-dependency-trends} shows the corresponding yearly trajectories. Determiners attached to prepositional-object heads became less frequent, while prenominal modification increased relative to prepositional modification. Together, these patterns are consistent with less prepositional elaboration and relatively more compact noun-phrase modification.

\input{figures/main-dependencies}
\input{figures/dependency-bigrams-changes}

\subsubsection{Directed Dependency-Edge Bigrams}

To represent relations between dependency roles, we defined each non-punctuation dependency edge as the ordered pair of its head and child labels, written \texttt{A$\rightarrow$B}. For example, \texttt{prep$\rightarrow$pobj} denotes an edge from a prepositional modifier to its object. We aggregated each bigram by year as a proportion of all non-punctuation dependency edges and estimated its post-2022 change in slope.

For display, we ranked bigrams by their mean absolute slope change across the four abstract corpora, giving each corpus equal weight, and retained the twelve largest. The strongest cross-corpus patterns included declines in \texttt{prep$\rightarrow$pobj} and \texttt{pobj$\rightarrow$det}, alongside increases in relations involving adjectival and compound modification. These patterns complement Fig.~\ref{fig:main-dependency-trends}, but do not establish direct substitution between constructions.

\section{Additional Analyses}

\subsection{Cross-Corpus Similarity of Linguistic-Change Profiles}
\label{app:cross-corpus-similarity}

To complement the feature-specific comparisons, we examined whether the
overall pattern of post-intervention trajectory changes was similar across
corpora. For each corpus pair, we correlated the standardized ITS slope-change
estimates across all jointly estimable features. This analysis captures
\emph{relative profile similarity}: a high correlation means that features
showing comparatively larger or more positive changes in one corpus tend to
do so in the other corpus as well. It does not imply that the absolute effect
magnitudes are equal, that the same features reach statistical significance,
or that the corpora share a causal mechanism.

The three arXiv corpora show highly similar feature-change profiles
($r=.93$--$.95$). Correlations between medRxiv and the arXiv corpora remain
positive but are lower ($r=.69$--$.74$), suggesting a more domain- or platform-specific pattern in the medical corpus.

\input{figures/corpus-correlation}

\subsection{Titles as a Secondary Text Scope} \label{app:title-abstract-comparison}
The main text reports directional agreement between title and abstract
trajectory changes. Table~\ref{tab:title-abstract-summary} provides complementary
information about feature coverage and differences in standardized effect
magnitude. Features that could not be estimated reliably are
retained as missing rather than interpreted as null effects. Most incomplete pairs in Table~\ref{tab:title-abstract-summary} occur because a feature cannot be estimated reliably in titles.

\input{figures/main-titles}
\input{tables/titles}

\subsection{Topic Composition and Topic-Level Trajectories}
\label{app:topic-trajectories}

We use the topic-stratified analyses to assess whether the corpus-level
linguistic changes could be explained primarily by shifts in the prevalence
of particular research topics. Figure~\ref{fig:topic-composition-four-corpora}
shows how the relative prevalence of corpus-specific BERTopic clusters
changed over time. Figure~\ref{fig:topic-example-trajectories} then shows
yearly trajectories for the five representative features highlighted in the
main text, estimated separately within each topic. Together, these figures
help assess changes in corpus composition from linguistic trajectories that
appear across multiple topics. Because topic models were fitted independently
within each corpus, topic identifiers and labels are corpus-specific and
should not be compared across corpora.

\input{figures/appendix-topic-composition}

\input{figures/appendix-topic-example-trajectories}

\FloatBarrier

\section{Temporal Placebo Checks}

\label{app:placebo-checks}

In Table~\ref{tab:placebo}, we refitted the interrupted-time-series models using placebo interventions in January 2018–2021; medRxiv supports only the 2021 placebo because of its shorter history. Earlier cutoffs often produced changes in the same direction, indicating pre-existing drift. However, the prespecified November 2022 intervention yielded the largest absolute effect in 19 of 20 feature–corpus comparisons. The results therefore suggest that changes in these trajectories were generally larger around the November 2022 breakpoint than at earlier placebo cutoffs, rather than originating entirely at that point. For a compact illustration, we show one feature from each prespecified group in the main results table, selecting the feature with the largest mean absolute primary effect across corpora. Selection did not use the placebo estimates, and the displayed subset is not a summary of all placebo tests.

\input{tables/placebo.tex}

\clearpage
\onecolumn
\begingroup
\let\vss\vfil
\section{Full Results} \label{app:supplementary-materials}

\subsection{Interrupted Time-Series Estimates}
\input{tables/its-full}

\clearpage
\subsection{Counterfactual-Excess Estimates}
\input{tables/excess}
\clearpage
\endgroup

\end{document}

%% file: figures/freq-examples.tex
\begin{figure*}[t]
    \centering

    \begin{subfigure}[t]{1\linewidth}
        \centering
        \includegraphics[width=\linewidth]{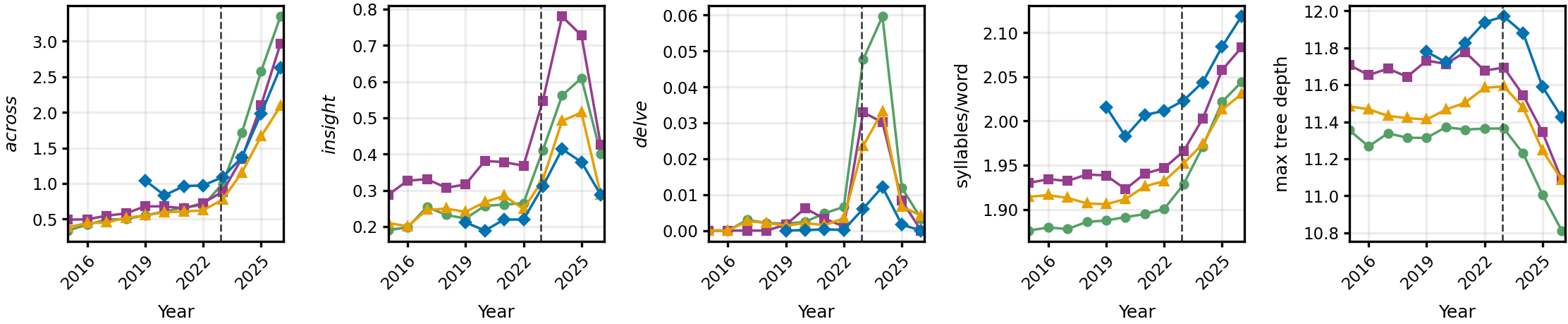}
    \end{subfigure}

    \begin{subfigure}[t]{0.37\linewidth}
        \centering
        \includegraphics[width=\linewidth]{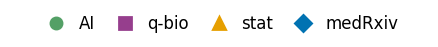}
    \end{subfigure}

    \caption{Selected yearly trajectories illustrating sustained increases
(\textit{across} and syllables per word), transient peaks (\textit{insight} and \textit{delve}), and trajectory reversal (max tree depth). Lexical frequencies are reported per 1,000 words;
the dashed line marks ChatGPT's release.}
    \label{fig:freq-examples}
\end{figure*}

%% file: tables/main.tex
\begin{table*}[t]

\centering
\resizebox{0.82\linewidth}{!}{%

\begin{tabular}{lllll}
\toprule
Feature & arXiv AI & arXiv q-bio & arXiv stat & medRxiv \\
\midrule
$\it{across}$ & 6.10$^{***}$ [5.51, 6.69] & 3.88$^{***}$ [3.02, 4.73] & 3.88$^{***}$ [3.41, 4.35] & 3.10$^{***}$ [2.30, 3.90] \\
$\it{align}$ & 3.14$^{***}$ [2.71, 3.57] & 1.51$^{***}$ [1.20, 1.82] & 1.92$^{***}$ [1.62, 2.22] & 2.71$^{***}$ [2.20, 3.22] \\
$\it{while}$ & 3.24$^{***}$ [2.67, 3.82] & 1.80$^{***}$ [1.47, 2.13] & 1.80$^{***}$ [1.42, 2.17] & 2.07$^{***}$ [1.69, 2.44] \\
$\it{exhibit}$ & 2.19$^{***}$ [1.44, 2.93] & 0.76$^{***}$ [0.52, 1.00] & 0.95$^{***}$ [0.64, 1.25] & 1.37$^{***}$ [0.75, 1.99] \\
\addlinespace[2pt]
em dash & 2.35$^{***}$ [1.45, 3.24] & 1.45$^{***}$ [0.82, 2.07] & 1.46$^{***}$ [0.76, 2.17] & 2.84$^{***}$ [2.15, 3.54] \\
\addlinespace[2pt]
syllables/word & 4.15$^{***}$ [3.90, 4.41] & 3.78$^{***}$ [3.48, 4.08] & 2.75$^{***}$ [2.48, 3.02] & 1.24$^{***}$ [0.67, 1.81] \\
FRE & -3.50$^{***}$ [-3.65, -3.34] & -3.41$^{***}$ [-3.70, -3.11] & -2.56$^{***}$ [-2.88, -2.24] & -0.87$^{***}$ [-1.32, -0.41] \\
Dale-Chall & 2.56$^{***}$ [2.42, 2.70] & 3.16$^{***}$ [2.87, 3.45] & 3.40$^{***}$ [3.03, 3.77] & 0.76$^{**}$ [0.20, 1.31] \\
Gunning Fog & 4.02$^{***}$ [3.82, 4.22] & 2.73$^{***}$ [2.46, 3.00] & 1.89$^{***}$ [1.52, 2.26] & 0.59$^{***}$ [0.25, 0.93] \\
\addlinespace[2pt]
max tree depth & -2.13$^{***}$ [-2.45, -1.82] & -1.31$^{***}$ [-1.61, -1.01] & -1.69$^{***}$ [-2.11, -1.27] & -1.88$^{***}$ [-2.25, -1.51] \\
tree depth $\sigma$ & -1.93$^{***}$ [-2.13, -1.72] & -1.12$^{***}$ [-1.31, -0.93] & -1.42$^{***}$ [-1.76, -1.08] & -1.91$^{***}$ [-2.29, -1.54] \\
list of three & 2.69$^{***}$ [2.09, 3.29] & 1.14$^{***}$ [0.78, 1.50] & 1.60$^{***}$ [1.00, 2.20] & 0.55$^{*}$ [0.10, 1.00] \\
coord. count & 0.57$^{***}$ [0.36, 0.79] & 0.35$^{**}$ [0.13, 0.57] & 0.92$^{***}$ [0.54, 1.31] & -0.59$^{***}$ [-0.91, -0.28] \\
\addlinespace[2pt]
$\it{moreover}$ & -1.54$^{***}$ [-1.77, -1.31] & -0.81$^{***}$ [-1.01, -0.61] & -1.12$^{***}$ [-1.43, -0.80] & -0.59$^{*}$ [-1.12, -0.06] \\
$\it{crucially}$ & 1.91$^{***}$ [1.03, 2.78] & 0.68$^{**}$ [0.29, 1.06] & 0.66$^{*}$ [0.12, 1.20] & 0.57$^{***}$ [0.31, 0.83] \\
$\it{thus}$ & -1.69$^{***}$ [-1.79, -1.58] & -0.68$^{***}$ [-0.81, -0.54] & -0.78$^{***}$ [-0.98, -0.58] & -0.64$^{**}$ [-1.01, -0.27] \\
hedging & -1.31$^{***}$ [-1.65, -0.96] & -0.56$^{***}$ [-0.74, -0.38] & -0.98$^{***}$ [-1.23, -0.73] & 0.81$^{***}$ [0.59, 1.02] \\
\bottomrule
\end{tabular}
}

\caption{Largest consistently significant post-intervention trajectory changes by feature group. Cells report standardized interrupted-time-series slope changes per year with 95\% confidence intervals. Within each prespecified group---lexical markers, punctuation, readability, syntax, and rhetorical stance---we report up to four individual features with the largest mean absolute standardized slope change across the four corpora. Eligible features required estimable effects and family-level FDR-adjusted \(q<.05\) in every corpus; aggregate group measures were excluded. \(^{*}q<.05\), \(^{**}q<.01\), and \(^{***}q<.001\), after Benjamini–Hochberg correction within each feature family and corpus.}
\label{table:main}
\end{table*}

%% file: figures/main-punctuation-forest.tex
\begin{figure}[h]
    \centering

    \begin{subfigure}[t]{0.9\linewidth}
        \centering
        \includegraphics[width=\linewidth]{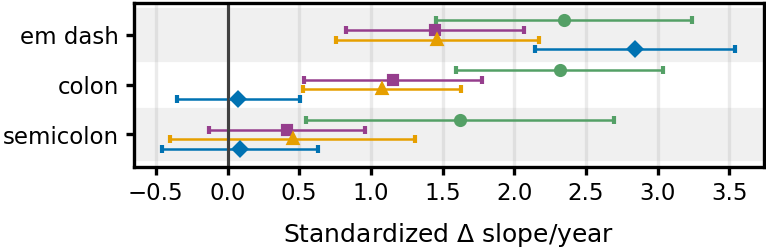}
    \end{subfigure}

    \begin{subfigure}[t]{0.69\linewidth}
        \centering
        \includegraphics[width=\linewidth]{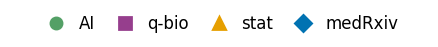}
    \end{subfigure}

    \caption{
Standardized post-intervention slope changes for punctuation features across the four abstract corpora, with 95\% confidence intervals.
    }
    \label{fig:main-punctuation-forest}
    
\end{figure}

%% file: figures/main-readability-forest.tex
\begin{figure}[h]
    \centering

    \begin{subfigure}[t]{0.9\linewidth}
        \centering
        \includegraphics[width=\linewidth]{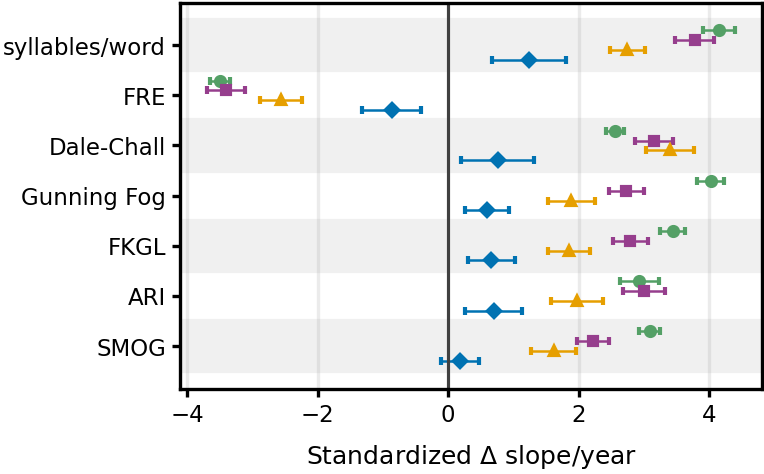}
    \end{subfigure}

    \begin{subfigure}[t]{0.69\linewidth}
        \centering
        \includegraphics[width=\linewidth]{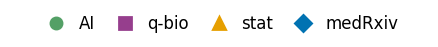}
    \end{subfigure}

    \caption{Standardized post-intervention slope changes for readability features across the four abstract corpora, with 95\% confidence intervals. Negative FRE changes indicate reduced reading ease.
    }
    \label{fig:main-readability-forest}
    
\end{figure}

%% file: figures/main-readability-decompose.tex
\begin{figure}[h]
    \centering

    \begin{subfigure}[t]{0.49\linewidth}
        \centering
        \includegraphics[width=\linewidth]{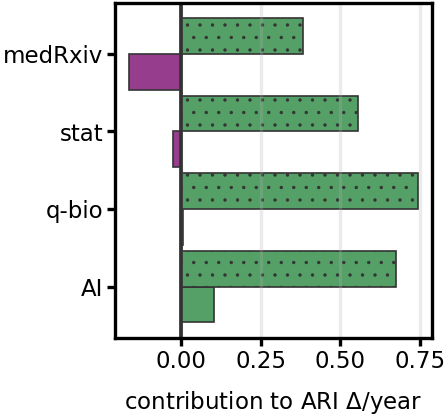}
    \end{subfigure}
    \hspace{-0.5em}
    \begin{subfigure}[t]{0.48\linewidth}
        \centering
        \includegraphics[width=\linewidth]{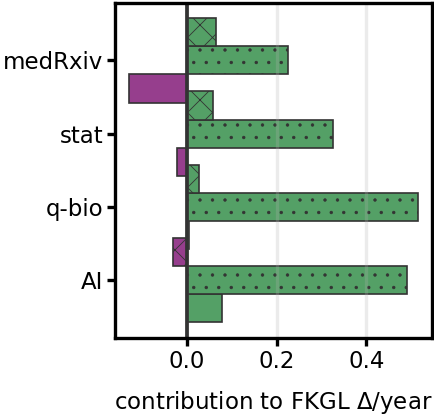}
    \end{subfigure}

    \begin{subfigure}[t]{0.49\linewidth}
        \centering
        \includegraphics[width=\linewidth]{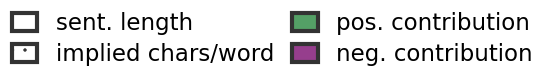}
    \end{subfigure}
    \hspace{-0.4em}
    \begin{subfigure}[t]{0.48\linewidth}
        \centering
        \includegraphics[width=\linewidth]{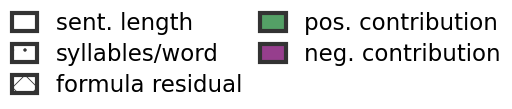}
    \end{subfigure}


    \caption{Decomposition of the annual post-intervention slope changes in ARI (left) and FKGL (right) into sentence- and word-length components across the four abstract corpora.
    }
    \label{fig:main-readability-decompose}
    
\end{figure}

%% file: figures/main-syntax-forest.tex
\begin{figure}[h]
    \centering

    \begin{subfigure}[t]{0.9\linewidth}
        \centering
        \includegraphics[width=\linewidth]{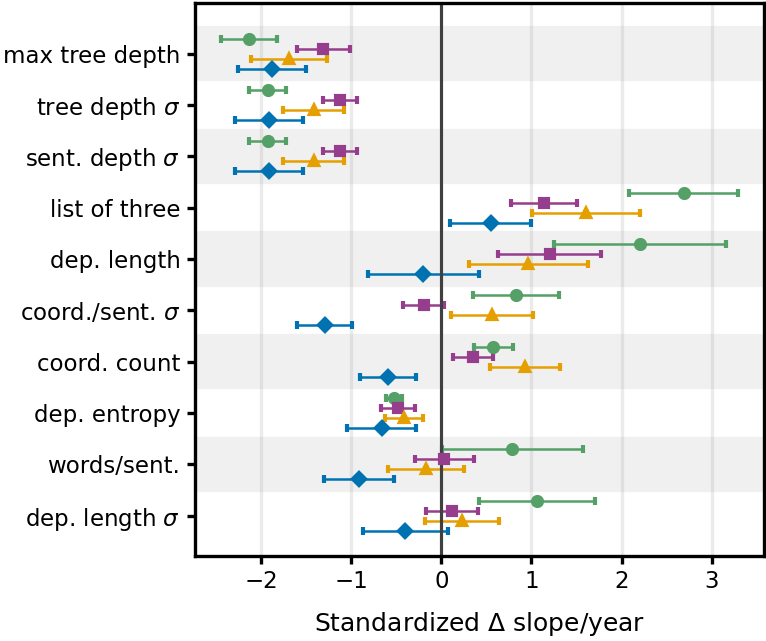}
    \end{subfigure}

    \begin{subfigure}[t]{0.69\linewidth}
        \centering
        \includegraphics[width=\linewidth]{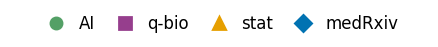}
    \end{subfigure}

    \caption{
Standardized post-intervention slope changes for syntactic features across the four abstract corpora, with 95\% confidence intervals.
    }
    \label{fig:main-syntax-forest}
    
\end{figure}

%% file: figures/main-dependency-trends.tex
\begin{figure}[h]
    \centering

    \begin{subfigure}[t]{0.99\linewidth}
        \centering
        \includegraphics[width=\linewidth]{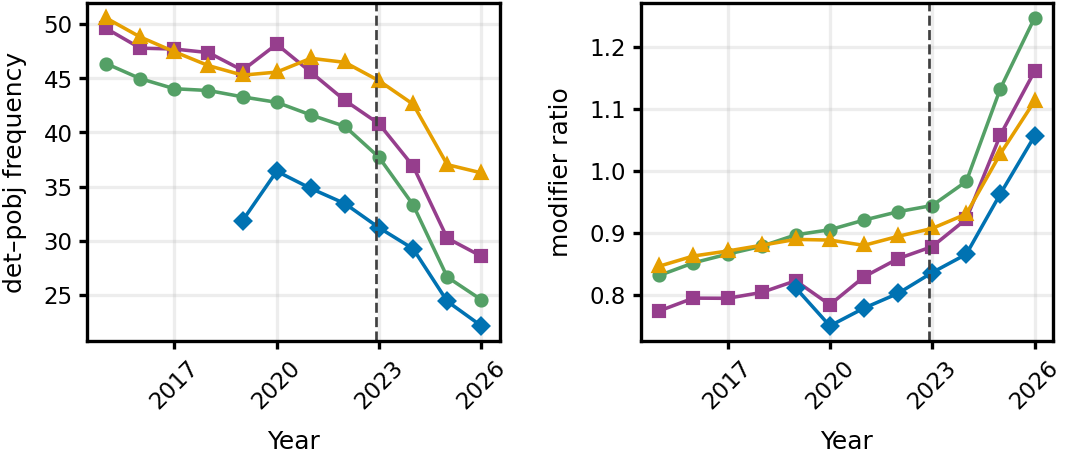}
    \end{subfigure}

    \begin{subfigure}[t]{0.69\linewidth}
        \centering
        \includegraphics[width=\linewidth]{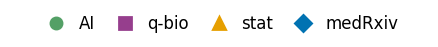}
    \end{subfigure}

    \caption{Dependency-context trajectories across the four abstract corpora. The left panel shows determiners attached to prepositional objects per 1,000 words. The right panel shows the ratio of adjectival and compound prenominal modifiers to prepositional modifiers. The dashed line marks the public release of ChatGPT.
    }
    \label{fig:main-dependency-trends}
    
\end{figure}

%% file: figures/main-topics.tex
\begin{figure*}[t]
    \centering

    \begin{subfigure}[c]{0.8\linewidth}
        \centering
        \includegraphics[width=\linewidth]{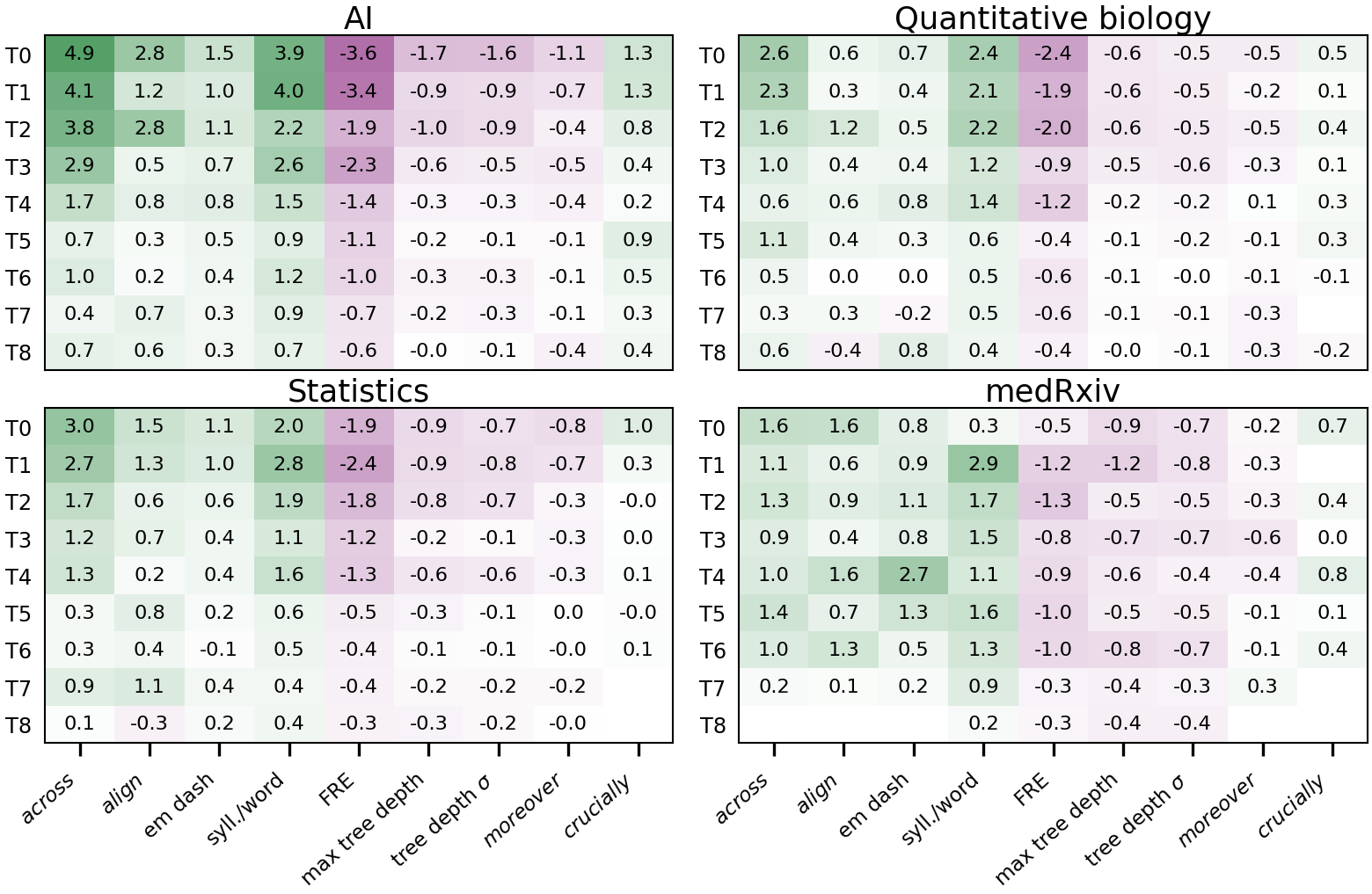}
    \end{subfigure}
    \begin{subfigure}[c]{0.09\linewidth}
        \centering
        \includegraphics[width=\linewidth]{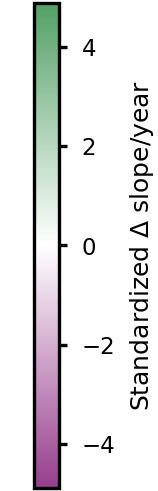}
    \end{subfigure}

    \caption{Topic-level heterogeneity in post-2022 abstract trajectories in arXiv (AI, quantitative biology, statistics) and medRxiv (medicine). Cells show standardized ITS slope changes for corpus-specific topics. We report up to the two largest individual-feature effects per prespecified group among features significant in all four corpora; aggregate measures are excluded. Green indicates positive and orchid negative slope changes. Topic identifiers are corpus-specific and are not matched across panels.
    }
    \label{fig:main-topics-heatmap}
    
\end{figure*}

%% file: tables/features.tex
\begin{table*}[t]
\centering
\small
\begin{tabularx}{\textwidth}{@{}lXl@{}}
\toprule
Feature(s) & Operational definition & Unit \\
\midrule

\multicolumn{3}{@{}l}{\textbf{Lexical and discourse features}} \\
Single-word markers 
    & Whole-token matches in lowercased lemmatized text 
    & per 1,000 words \\
Multiword phrases 
    & Whole-phrase matches in lowercased normalized text 
    & per 1,000 words \\
Punctuation 
    & Counts of colons, semicolons, and normalized em dashes 
    & per 1,000 words \\
Certainty / hedging 
    & Token matches against the corresponding lemmatized lexicons 
    & proportion \\

\addlinespace
\multicolumn{3}{@{}l}{\textbf{Readability features}} \\
Words / sentence
    & Number of alphabetic spaCy tokens divided by the sentence count
    & words / sentence \\
Syllables / word
    & Total \texttt{textstat} syllable count divided by the number of alphabetic spaCy tokens
    & syllables / word \\
Readability scores 
    & FRE, FKGL, Dale--Chall, SMOG, ARI, Gunning Fog 
    & native score \\

\addlinespace
\multicolumn{3}{@{}l}{\textbf{Syntactic features}} \\
Dependency-tree depth 
    & Maximum root-to-terminal-token path over sentence dependency trees 
    & tokens \\
Tree-depth variation
    & Sample SD of sentence-level dependency-tree depths; zero for fewer than two sentences
    & tokens \\
Dependency length
    & Mean absolute token--head index distance over non-root arcs; punctuation retained
    & tokens \\
Dependency-length variation
    & Sample SD of sentence-level mean dependency lengths; zero when fewer than two sentence means are available
    & tokens \\
Dependency-label entropy
    & Shannon entropy, using the natural logarithm, over dependency labels; \texttt{punct} excluded and \texttt{ROOT} retained
    & nats \\
Coordination
    & Number of \texttt{cc} tokens; variation is the sample SD of sentence-level counts
    & count \\
List of three 
    & \texttt{and}/\texttt{or} coordination with exactly three conjuncts 
    & per 1,000 words \\

\bottomrule
\end{tabularx}

\caption{
Operational definitions of the analyzed linguistic features. Features were computed at document level and aggregated as monthly arithmetic means; monthly paper counts were used as weights in the inferential models. Frequency features were normalized as $1000c/(N_{\alpha}+1)$ and certainty and hedging as $c/(N_{\alpha}+1)$, where $c$ is the document-level count and $N_{\alpha}$ is the number of alphabetic spaCy tokens.
}
\label{tab:features}
\end{table*}

%% file: figures/main-dependencies.tex
\begin{figure}[h]
    \centering

    \begin{subfigure}[t]{0.59\linewidth}
        \centering
        \includegraphics[width=\linewidth]{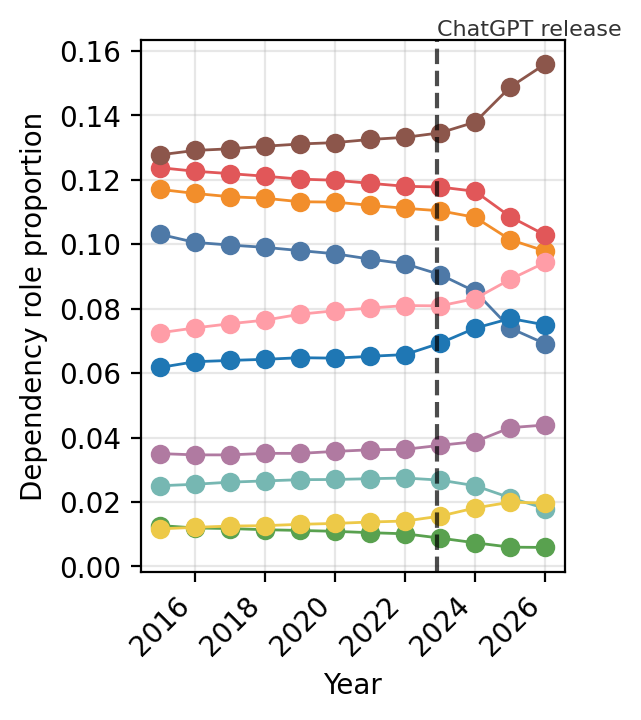}
    \end{subfigure}
    \begin{subfigure}[t]{0.39\linewidth}
        \centering
        \includegraphics[width=\linewidth]{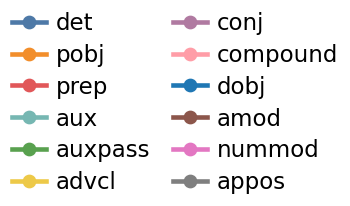}
    \end{subfigure}

    \caption{
Yearly dependency-role proportions in arXiv AI abstracts for the roles with the largest absolute post-2022 changes in slope.
    }
    \label{fig:main-dependencies}
    
\end{figure}

%% file: figures/dependency-bigrams-changes.tex
\begin{figure}[h]
    \centering

    \begin{subfigure}[t]{0.93\linewidth}
        \centering
        \includegraphics[width=\linewidth]{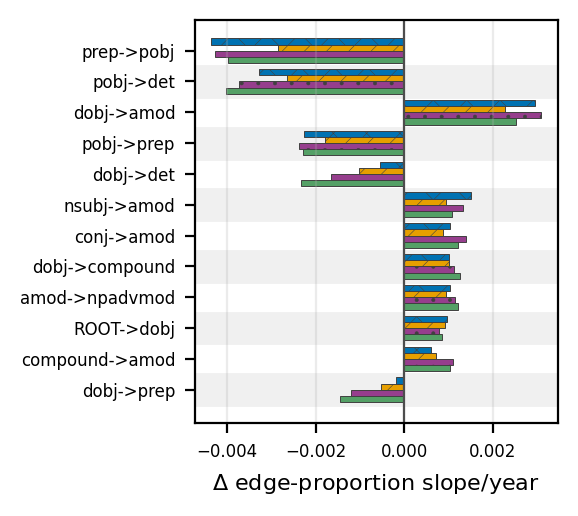}
    \end{subfigure}

    \begin{subfigure}[t]{0.7\linewidth}
        \centering
        \includegraphics[width=\linewidth]{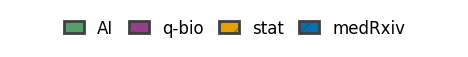}
    \end{subfigure}

\caption{
Post-2022 changes in the annual slopes of directed dependency-edge proportions across the four abstract corpora. A bigram \texttt{A$\rightarrow$B} denotes an edge from a head to its child. Bars report the change in slope after 2022 relative to the pre-2022 trajectory. Positive values indicate a more positive post-2022 trend; negative values indicate a more negative trend.
}
    \label{fig:dependency-bigram-slopes}
    
\end{figure}

%% file: figures/corpus-correlation.tex
\begin{figure}[h]
    \centering

    \begin{subfigure}[t]{0.75\linewidth}
        \centering
        \includegraphics[width=\linewidth]{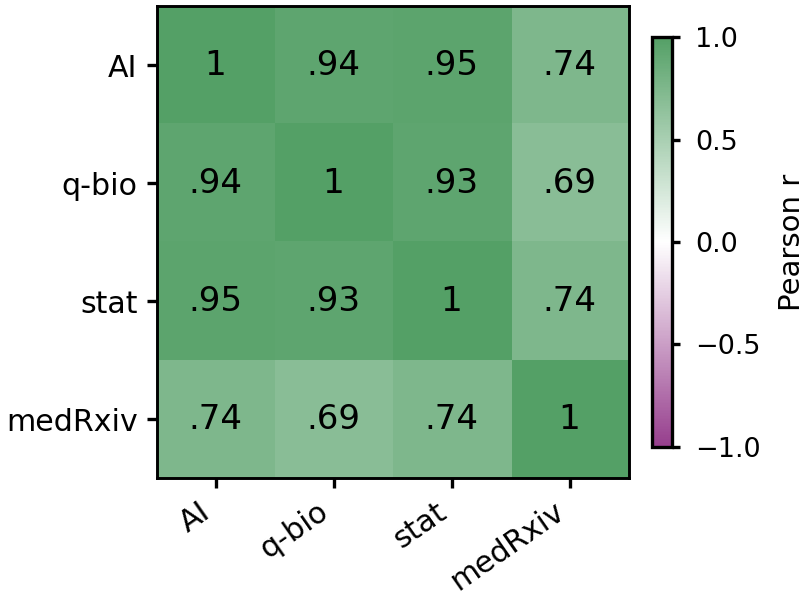}
    \end{subfigure}

    \caption{
    Pairwise similarity of post-2022 linguistic-change profiles across the abstract corpora. Cells report Pearson correlations between standardized ITS slope-change estimates across all features jointly estimable in each corpus pair.
    }
    \label{fig:corpus-correlation-heatmap}
    
\end{figure}

%% file: figures/main-titles.tex
\begin{figure}[h]
    \centering

    \begin{subfigure}[t]{0.9\linewidth}
        \centering
        \includegraphics[width=\linewidth]{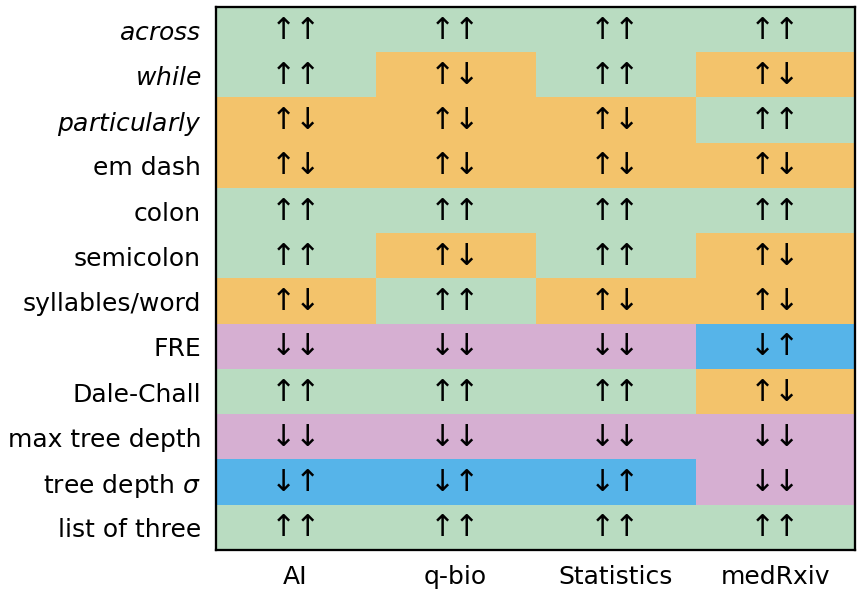}
    \end{subfigure}

    \begin{subfigure}[t]{0.59\linewidth}
        \centering
        \includegraphics[width=\linewidth]{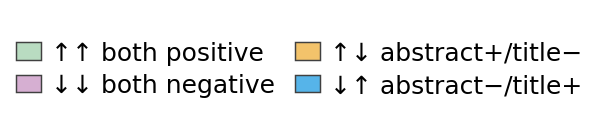}
    \end{subfigure}

    \caption{
Directional agreement between abstract and title trajectories. Arrows show the standardized ITS slope-change direction for abstracts and titles. In each cell, the first arrow indicates the standardized ITS slope-change direction for abstracts and the second for titles.
    }
    \label{fig:main-titles}
    
\end{figure}

%% file: tables/titles.tex
\begin{table}[h]
\centering
\small
\resizebox{\columnwidth}{!}{
\begin{tabular}{@{}lrrrr@{}}
\toprule
 & AI & q-bio & stat & medRxiv \\
\midrule
Complete pairs
    & 57 & 53 & 50 & 55 \\

Inestimable in titles
    & 29 & 33 & 36 & 31 \\

Profile correlation, $r$
    & .49 & .59 & .34 & .45 \\

Median $|\Delta_{\mathrm{title}}
-\Delta_{\mathrm{abstract}}|$
    & 1.60 & 0.75 & 1.05 & 0.90 \\

Larger title effect
    & 3.5\% & 5.7\% & 4.0\% & 23.6\% \\
\bottomrule
\end{tabular}
}
\caption{
Complementary title--abstract comparison across 86 configured features. Complete pairs are features with estimable standardized ITS slope changes in both abstract and title scopes. The median difference is measured on the standardized slope-change scale. \emph{Larger title effect} is the percentage of complete feature pairs for which the absolute standardized ITS slope change is larger in titles
than in abstracts.
}
\label{tab:title-abstract-summary}
\end{table}

%% file: figures/appendix-topic-composition.tex
\begin{figure*}[h]
\centering

\begin{subfigure}[t]{0.35\textwidth}
    \centering
    \includegraphics[
        width=0.88\linewidth,
        trim=5 5 5 5,
        clip
    ]{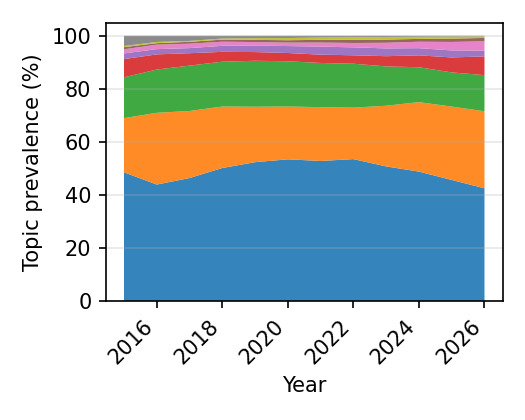}
    \caption{arXiv artificial intelligence}
    \label{fig:topic-composition-ai}
\end{subfigure}
\hfill
\begin{subfigure}[t]{0.35\textwidth}
    \centering
    \includegraphics[
        width=0.88\linewidth,
        trim=5 5 5 5,
        clip
    ]{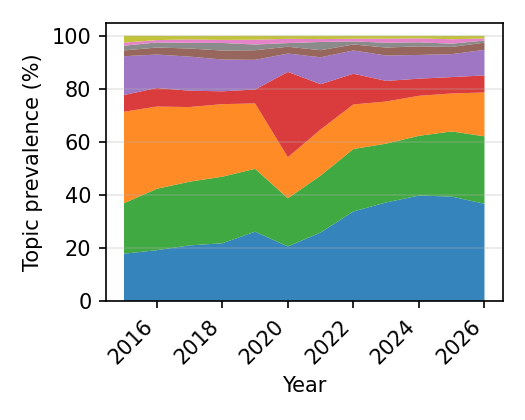}
    \caption{arXiv quantitative biology}
    \label{fig:topic-composition-qbio}
\end{subfigure}

\medskip

\begin{subfigure}[t]{0.35\textwidth}
    \centering
    \includegraphics[
        width=0.88\linewidth,
        trim=5 5 5 5,
        clip
    ]{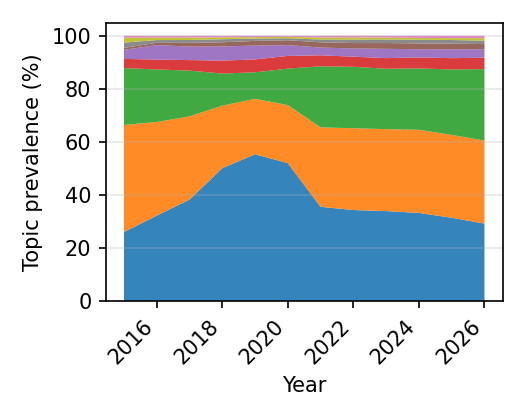}
    \caption{arXiv statistics}
    \label{fig:topic-composition-stat}
\end{subfigure}
\hfill
\begin{subfigure}[t]{0.35\textwidth}
    \centering
    \includegraphics[
        width=0.88\linewidth,
        trim=5 5 5 5,
        clip
    ]{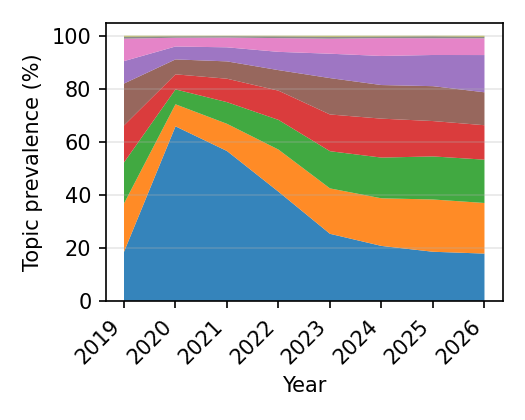}
    \caption{medRxiv}
    \label{fig:topic-composition-medrxiv}
\end{subfigure}

\medskip

\begin{subfigure}[t]{\textwidth}
    \centering
    \includegraphics[
        width=\linewidth,
        trim=2 2 2 2,
        clip
    ]{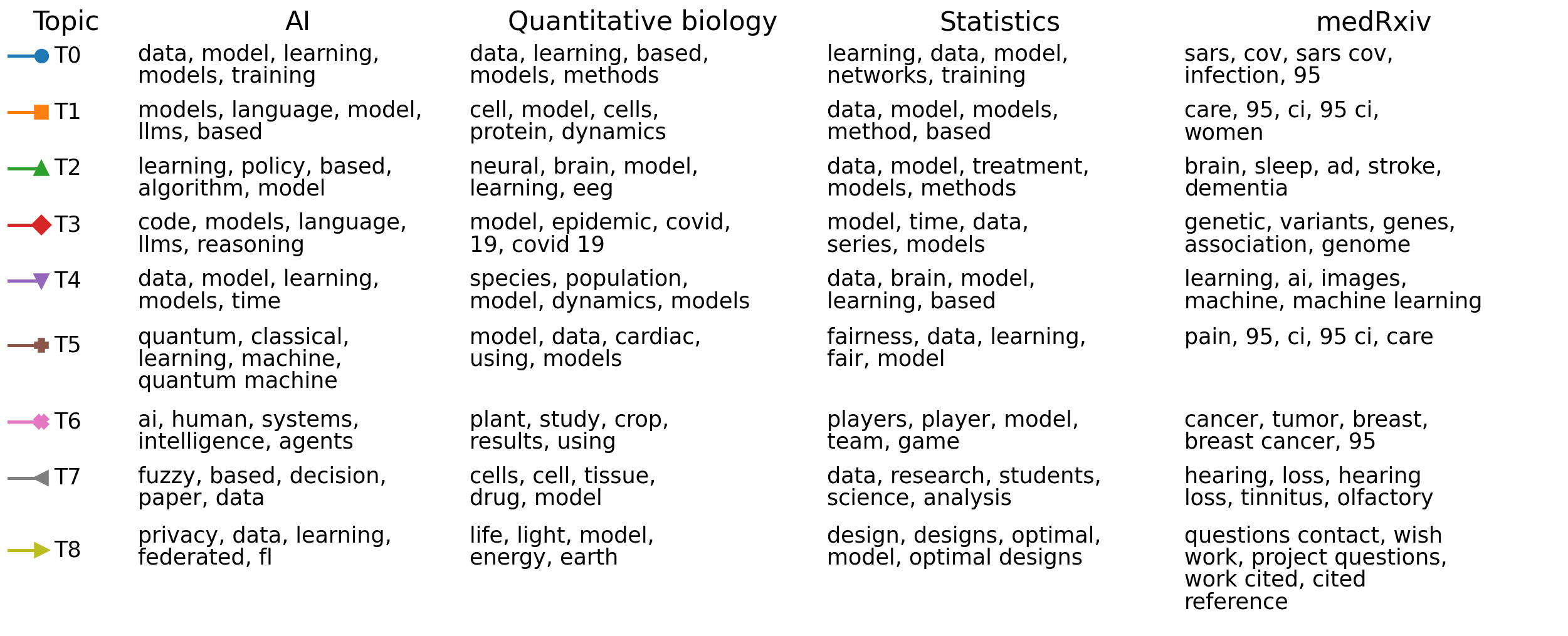}
\end{subfigure}

\caption{
    Topic composition over time in the four abstract corpora.
    Stacked areas show the annual prevalence of corpus-specific
    BERTopic clusters. The combined legend reports topic identifiers
    and shortened c-TF-IDF labels separately for each corpus.
    Topic identifiers are corpus-specific and are not comparable
    across panels. Values for 2026 include records through April.
}
\label{fig:topic-composition-four-corpora}

\end{figure*}

%% file: figures/appendix-topic-example-trajectories.tex
\begin{figure*}[t]
    \centering

    \begin{subfigure}[t]{\linewidth}
        \centering
        \includegraphics[width=\linewidth]{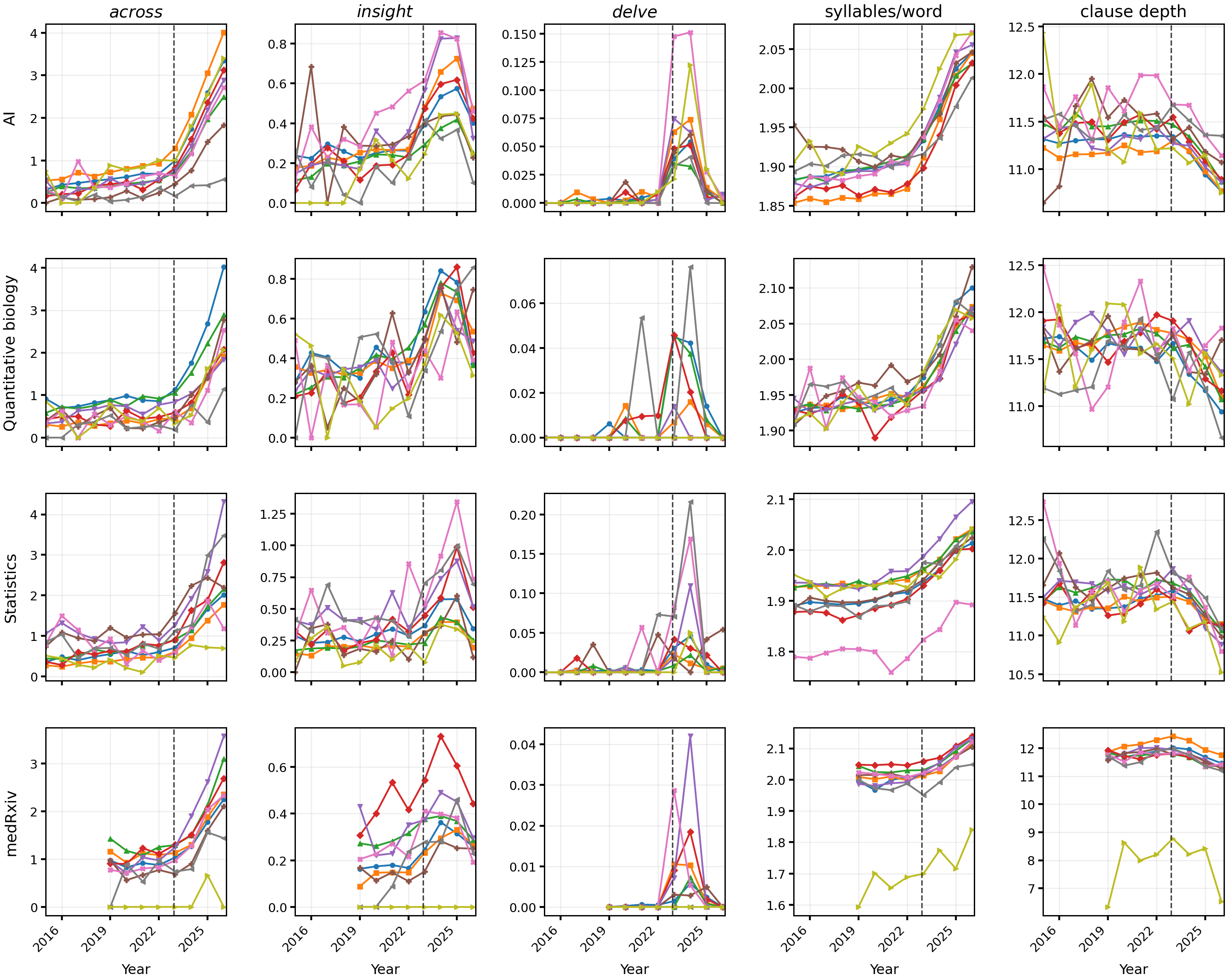}
    \end{subfigure}

    \vspace{0.3em}

    \begin{subfigure}[t]{\linewidth}
        \centering
        \includegraphics[width=\linewidth]{assets/topic_example_trajectories_legend.png}
    \end{subfigure}

    \caption{Topic-level yearly trajectories for the five representative features shown in Figure~\ref{fig:freq-examples}. Rows correspond to
    the four abstract corpora and lines to corpus-specific BERTopic clusters. Lexical frequencies are reported per 1,000 words; the dashed line marks ChatGPT's release. \textit{Clause depths }stands for maximum tree depth. Topic identifiers are corpus-specific and are not comparable across rows.}
    \label{fig:topic-example-trajectories}
\end{figure*}

%% file: tables/placebo.tex
\begin{table}[h]
\centering
\small
\resizebox{\columnwidth}{!}{%
\begin{tabular}{@{}llrrrrr@{}}
\toprule
Corpus & Feature & 2018 & 2019 & 2020 & 2021 & Actual \\
\midrule
AI & $\it{across}$ & 2.98 & 3.65 & 4.19 & 4.80 & \textbf{6.10} \\
 & em dash & 0.47 & 0.81 & 0.96 & 1.18 & \textbf{2.35} \\
 & syll./word & 2.80 & 2.77 & 3.25 & 3.75 & \textbf{4.15} \\
 & max tree depth & -0.51 & -0.58 & -0.80 & -1.20 & \textbf{-2.13} \\
 & $\it{moreover}$ & -0.17 & -0.21 & -0.48 & -0.85 & \textbf{-1.54} \\
\addlinespace[3pt]
q-bio & $\it{across}$ & 1.39 & 1.45 & 1.63 & 2.30 & \textbf{3.88} \\
 & em dash & 0.25 & 0.33 & 0.40 & 0.62 & \textbf{1.45} \\
 & syll./word & 1.87 & 1.96 & 2.44 & 2.81 & \textbf{3.78} \\
 & max tree depth & -0.27 & -0.38 & -0.60 & -0.84 & \textbf{-1.31} \\
 & $\it{moreover}$ & -0.32 & -0.37 & -0.37 & -0.50 & \textbf{-0.81} \\
\addlinespace[3pt]
stat & $\it{across}$ & 1.28 & 1.40 & 1.71 & 2.38 & \textbf{3.88} \\
 & em dash & 0.17 & 0.43 & 0.61 & 0.77 & \textbf{1.46} \\
 & syll./word & 2.23 & 2.71 & 2.92 & \textbf{3.20} & 2.75 \\
 & max tree depth & -0.06 & -0.06 & -0.25 & -0.78 & \textbf{-1.69} \\
 & $\it{moreover}$ & -0.49 & -0.28 & -0.40 & -0.66 & \textbf{-1.12} \\
\addlinespace[3pt]
medRxiv & $\it{across}$ & -- & -- & -- & 1.85 & \textbf{3.10} \\
 & em dash & -- & -- & -- & 1.67 & \textbf{2.84} \\
 & syll./word & -- & -- & -- & 0.79 & \textbf{1.24} \\
 & max tree depth & -- & -- & -- & -1.55 & \textbf{-1.88} \\
 & $\it{moreover}$ & -- & -- & -- & 0.38 & \textbf{-0.59} \\
\bottomrule
\end{tabular}
}
\caption{Temporal placebo checks for selected main-result features. Cells report standardized annual ITS slope changes for January placebo interventions and the prespecified November 2022 intervention (Actual, in bold). Dashes indicate unavailable estimates.}
\label{tab:placebo}
\end{table}

%% file: tables/its-full.tex
\begingroup
\small
\setlength{\tabcolsep}{3.5pt}
\begin{longtable}{@{}lllll@{}}
\toprule
Feature & AI & q-bio & stat & medRxiv \\
\midrule
\endfirsthead
\toprule
Feature & AI & q-bio & stat & medRxiv \\
\midrule
\endhead
\multicolumn{5}{l}{\textbf{Adjectives}} \\
$\it{comprehensive}$ & -0.20 [-1.83, 1.43] & -0.07 [-1.30, 1.16] & -0.07 [-0.87, 0.73] & 0.15 [-0.61, 0.91] \\
$\it{crucial}$ & -1.30 [-2.92, 0.32] & -0.60 [-1.83, 0.62] & -0.67 [-1.72, 0.39] & 0.16 [-1.01, 1.33] \\
$\it{intricate}$ & -1.26 [-3.96, 1.44] & -0.71 [-2.31, 0.88] & -0.45 [-1.93, 1.02] & -0.45 [-1.94, 1.04] \\
$\it{pivotal}$ & -1.17 [-3.98, 1.65] & -0.83 [-2.31, 0.65] & -0.24 [-0.86, 0.38] & -0.01 [-0.24, 0.21] \\
$\it{potential}$ & -1.39 [-3.18, 0.41] & -0.43 [-1.21, 0.36] & -0.21 [-0.92, 0.51] & 0.08 [-0.71, 0.86] \\
\addlinespace[2pt]
\multicolumn{5}{l}{\textbf{Causal markers}} \\
$\it{accordingly}$ & -0.43$^{***}$ [-0.58, -0.28] & -0.06 [-0.24, 0.12] & -0.18 [-0.38, 0.01] & 0.13$^{**}$ [0.04, 0.21] \\
$\it{consequently}$ & -0.66$^{***}$ [-1.03, -0.30] & -0.32 [-0.62, -0.02] & -0.15 [-0.40, 0.10] & -0.44 [-0.88, 0.01] \\
$\it{hence}$ & -1.26$^{***}$ [-1.42, -1.09] & -0.64$^{***}$ [-0.86, -0.43] & -0.61$^{***}$ [-0.73, -0.49] & -0.28$^{*}$ [-0.50, -0.07] \\
$\it{thereby}$ & 1.30$^{*}$ [0.09, 2.51] & 0.19 [-0.28, 0.65] & 1.79$^{***}$ [1.41, 2.16] & 0.09 [-0.04, 0.23] \\
$\it{thus}$ & -1.69$^{***}$ [-1.79, -1.58] & -0.68$^{***}$ [-0.81, -0.54] & -0.78$^{***}$ [-0.98, -0.58] & -0.64$^{**}$ [-1.01, -0.27] \\
\addlinespace[2pt]
\multicolumn{5}{l}{\textbf{Contrast markers}} \\
$\it{alternatively}$ & -0.29$^{***}$ [-0.37, -0.21] & -0.20 [-0.47, 0.08] & -0.19 [-0.39, 0.02] & 0.07 [-0.27, 0.41] \\
$\it{conversely}$ & 0.34$^{**}$ [0.14, 0.55] & 0.02 [-0.26, 0.30] & 0.01 [-0.39, 0.41] & -0.07 [-0.50, 0.36] \\
$\it{however}$ & -1.02$^{***}$ [-1.37, -0.67] & -0.74$^{**}$ [-1.14, -0.34] & -0.74$^{**}$ [-1.26, -0.22] & -0.76$^{***}$ [-1.13, -0.39] \\
$\it{nevertheless}$ & -1.13$^{***}$ [-1.40, -0.87] & -0.29$^{**}$ [-0.45, -0.13] & -0.65$^{***}$ [-0.81, -0.50] & -0.27 [-0.50, -0.03] \\
$\it{nonetheless}$ & -0.86$^{***}$ [-1.30, -0.42] & -0.07 [-0.26, 0.11] & -0.31$^{**}$ [-0.53, -0.09] & -0.09 [-0.34, 0.16] \\
\addlinespace[2pt]
\multicolumn{5}{l}{\textbf{Emphasis markers}} \\
$\it{crucially}$ & 1.91$^{***}$ [1.03, 2.78] & 0.68$^{**}$ [0.29, 1.06] & 0.66$^{*}$ [0.12, 1.20] & 0.57$^{***}$ [0.31, 0.83] \\
$\it{importantly}$ & -0.29$^{***}$ [-0.41, -0.17] & -0.09 [-0.29, 0.11] & -0.35$^{**}$ [-0.56, -0.13] & 0.24$^{**}$ [0.07, 0.41] \\
$\it{notably}$ & 0.45 [-1.24, 2.14] & 0.20 [-0.59, 0.98] & -0.14 [-1.19, 0.91] & 0.86 [0.01, 1.71] \\
$\it{particularly}$ & 1.62 [-0.29, 3.52] & 0.82$^{*}$ [0.23, 1.41] & 1.47$^{*}$ [0.38, 2.56] & 2.38$^{***}$ [1.43, 3.33] \\
$\it{remarkably}$ & -0.34$^{**}$ [-0.58, -0.10] & -0.19 [-0.42, 0.05] & -0.22 [-0.55, 0.11] & -0.24 [-0.60, 0.12] \\
\addlinespace[2pt]
\multicolumn{5}{l}{\textbf{Marker words}} \\
$\it{across}$ & 6.10$^{***}$ [5.51, 6.69] & 3.88$^{***}$ [3.02, 4.73] & 3.88$^{***}$ [3.41, 4.35] & 3.10$^{***}$ [2.30, 3.90] \\
$\it{additionally}$ & -1.60 [-4.06, 0.85] & -0.60 [-1.99, 0.80] & -0.84 [-2.32, 0.65] & -1.00 [-2.15, 0.14] \\
$\it{finding}$ & 1.84$^{***}$ [0.97, 2.71] & 1.17$^{***}$ [0.63, 1.70] & 0.60 [-0.23, 1.42] & 2.02$^{***}$ [1.61, 2.42] \\
$\it{insight}$ & 0.54 [-0.89, 1.98] & 0.16 [-1.06, 1.39] & 0.39 [-0.92, 1.71] & -0.11 [-0.86, 0.63] \\
$\it{interplay}$ & 0.03 [-0.22, 0.28] & 0.19 [-0.02, 0.40] & -0.06 [-0.43, 0.30] & 0.14 [-0.34, 0.62] \\
$\it{meticulously}$ & -1.33 [-5.45, 2.79] & -1.01 [-1.98, -0.04] & -0.60 [-1.43, 0.22] & 0.10 [-0.67, 0.86] \\
$\it{notably}$ & 0.45 [-1.24, 2.14] & 0.20 [-0.59, 0.98] & -0.14 [-1.19, 0.91] & 0.86 [0.01, 1.71] \\
$\it{particularly}$ & 1.62 [-0.29, 3.52] & 0.82$^{*}$ [0.23, 1.41] & 1.47$^{*}$ [0.38, 2.56] & 2.38$^{***}$ [1.43, 3.33] \\
$\it{realm}$ & -1.90 [-3.63, -0.17] & -0.57 [-1.25, 0.10] & -0.72 [-1.60, 0.16] & -0.27 [-0.65, 0.10] \\
$\it{while}$ & 3.24$^{***}$ [2.67, 3.82] & 1.80$^{***}$ [1.47, 2.13] & 1.80$^{***}$ [1.42, 2.17] & 2.07$^{***}$ [1.69, 2.44] \\
$\it{within}$ & 0.89 [-0.26, 2.04] & 0.52$^{**}$ [0.19, 0.85] & 1.05$^{***}$ [0.58, 1.52] & 0.22 [-0.12, 0.55] \\
\addlinespace[2pt]
\multicolumn{5}{l}{\textbf{Other}} \\
certainty terms & 1.10 [-0.08, 2.29] & 0.65 [-0.02, 1.32] & 0.70$^{*}$ [0.15, 1.26] & 1.39$^{***}$ [1.01, 1.77] \\
hedge terms & -1.31$^{***}$ [-1.65, -0.96] & -0.56$^{***}$ [-0.74, -0.38] & -0.98$^{***}$ [-1.23, -0.73] & 0.81$^{***}$ [0.59, 1.02] \\
\addlinespace[2pt]
\multicolumn{5}{l}{\textbf{Phrases}} \\
$\it{comprehensive\ chapter}$ & -- & -- & -- & -- \\
$\it{deep\ dive}$ & -0.52$^{*}$ [-0.89, -0.15] & -0.23 [-0.55, 0.09] & 0.13 [-0.04, 0.29] & 0.31 [-0.04, 0.66] \\
$\it{essential\ insight}$ & -- & -- & -- & -- \\
$\it{intricate\ interplay}$ & -0.02 [-0.40, 0.35] & -0.13 [-0.56, 0.30] & 0.24 [-1.56, 2.04] & -1.86 [-5.52, 1.80] \\
$\it{intricate\ web}$ & -- & 0.02 [-0.02, 0.07] & -- & -0.01 [-0.19, 0.16] \\
$\it{meticulously\ delve}$ & -- & -- & -- & -- \\
\addlinespace[2pt]
\multicolumn{5}{l}{\textbf{Punctuation}} \\
colon & 2.32$^{***}$ [1.59, 3.04] & 1.15$^{***}$ [0.53, 1.77] & 1.08$^{***}$ [0.52, 1.63] & 0.08 [-0.35, 0.50] \\
em dash & 2.35$^{***}$ [1.45, 3.24] & 1.45$^{***}$ [0.82, 2.07] & 1.46$^{***}$ [0.76, 2.17] & 2.84$^{***}$ [2.15, 3.54] \\
semicolon & 1.62$^{**}$ [0.55, 2.70] & 0.41 [-0.13, 0.96] & 0.45 [-0.40, 1.31] & 0.09 [-0.45, 0.63] \\
\addlinespace[2pt]
\multicolumn{5}{l}{\textbf{Readability}} \\
ARI & 2.93$^{***}$ [2.64, 3.23] & 3.00$^{***}$ [2.69, 3.32] & 1.97$^{***}$ [1.57, 2.37] & 0.70$^{**}$ [0.25, 1.14] \\
syllables/word & 4.15$^{***}$ [3.90, 4.41] & 3.78$^{***}$ [3.48, 4.08] & 2.75$^{***}$ [2.48, 3.02] & 1.24$^{***}$ [0.67, 1.81] \\
Dale-Chall & 2.56$^{***}$ [2.42, 2.70] & 3.16$^{***}$ [2.87, 3.45] & 3.40$^{***}$ [3.03, 3.77] & 0.76$^{**}$ [0.20, 1.31] \\
FKGL & 3.44$^{***}$ [3.24, 3.64] & 2.79$^{***}$ [2.53, 3.06] & 1.85$^{***}$ [1.53, 2.17] & 0.66$^{***}$ [0.31, 1.02] \\
FRE & -3.50$^{***}$ [-3.65, -3.34] & -3.41$^{***}$ [-3.70, -3.11] & -2.56$^{***}$ [-2.88, -2.24] & -0.87$^{***}$ [-1.32, -0.41] \\
Gunning Fog & 4.02$^{***}$ [3.82, 4.22] & 2.73$^{***}$ [2.46, 3.00] & 1.89$^{***}$ [1.52, 2.26] & 0.59$^{***}$ [0.25, 0.93] \\
SMOG & 3.09$^{***}$ [2.93, 3.25] & 2.22$^{***}$ [1.98, 2.47] & 1.62$^{***}$ [1.27, 1.96] & 0.18 [-0.12, 0.47] \\
\addlinespace[2pt]
\multicolumn{5}{l}{\textbf{Sequential markers}} \\
$\it{additionally}$ & -1.60 [-4.06, 0.85] & -0.60 [-1.99, 0.80] & -0.84 [-2.32, 0.65] & -1.00 [-2.15, 0.14] \\
$\it{further}$ & -1.10$^{**}$ [-1.77, -0.42] & -0.59$^{***}$ [-0.82, -0.35] & -0.39$^{**}$ [-0.61, -0.17] & -0.42$^{*}$ [-0.80, -0.05] \\
$\it{furthermore}$ & -0.87$^{**}$ [-1.48, -0.26] & -0.53 [-1.01, -0.05] & -0.76$^{*}$ [-1.36, -0.17] & -1.06$^{***}$ [-1.48, -0.63] \\
$\it{moreover}$ & -1.54$^{***}$ [-1.77, -1.31] & -0.81$^{***}$ [-1.01, -0.61] & -1.12$^{***}$ [-1.43, -0.80] & -0.59$^{*}$ [-1.12, -0.06] \\
$\it{subsequently}$ & -0.97 [-2.08, 0.15] & -0.29 [-0.58, 0.00] & -0.48 [-0.99, 0.03] & -0.33$^{*}$ [-0.65, -0.02] \\
\addlinespace[2pt]
\multicolumn{5}{l}{\textbf{Summary markers}} \\
$\it{collectively}$ & 0.30$^{***}$ [0.16, 0.45] & 0.51$^{**}$ [0.23, 0.79] & -0.11 [-0.34, 0.12] & 0.35 [-0.07, 0.76] \\
$\it{in\ summary}$ & -0.25$^{**}$ [-0.40, -0.10] & -0.24 [-0.49, 0.00] & -0.18 [-0.41, 0.06] & -0.72$^{**}$ [-1.10, -0.34] \\
$\it{overall}$ & 0.44$^{*}$ [0.11, 0.77] & 0.16 [-0.10, 0.41] & 0.53$^{**}$ [0.21, 0.85] & -0.23 [-0.54, 0.08] \\
$\it{taken\ together}$ & 0.69$^{***}$ [0.42, 0.95] & 0.06 [-0.12, 0.23] & 0.46$^{**}$ [0.18, 0.74] & -0.37$^{**}$ [-0.59, -0.16] \\
$\it{ultimately}$ & 0.15 [-0.29, 0.59] & 0.18 [0.02, 0.34] & -0.12 [-0.51, 0.27] & -0.05 [-0.47, 0.37] \\
\addlinespace[2pt]
\multicolumn{5}{l}{\textbf{Syntax}} \\
words/sent. & 0.79$^{*}$ [0.01, 1.57] & 0.03 [-0.29, 0.36] & -0.17 [-0.59, 0.25] & -0.92$^{***}$ [-1.31, -0.53] \\
max tree depth & -2.13$^{***}$ [-2.45, -1.82] & -1.31$^{***}$ [-1.61, -1.01] & -1.69$^{***}$ [-2.11, -1.27] & -1.88$^{***}$ [-2.25, -1.51] \\
tree depth $\sigma$ & -1.93$^{***}$ [-2.13, -1.72] & -1.12$^{***}$ [-1.31, -0.93] & -1.42$^{***}$ [-1.76, -1.08] & -1.91$^{***}$ [-2.29, -1.54] \\
coord. count & 0.57$^{***}$ [0.36, 0.79] & 0.35$^{**}$ [0.13, 0.57] & 0.92$^{***}$ [0.54, 1.31] & -0.59$^{***}$ [-0.91, -0.28] \\
coord./sent. $\sigma$ & 0.83$^{***}$ [0.36, 1.30] & -0.20 [-0.42, 0.03] & 0.56$^{*}$ [0.11, 1.02] & -1.29$^{***}$ [-1.60, -0.99] \\
dep. entropy & -0.53$^{***}$ [-0.62, -0.44] & -0.48$^{***}$ [-0.67, -0.29] & -0.41$^{***}$ [-0.62, -0.20] & -0.66$^{**}$ [-1.04, -0.27] \\
dep. length & 2.20$^{***}$ [1.25, 3.16] & 1.20$^{***}$ [0.63, 1.77] & 0.96$^{**}$ [0.31, 1.62] & -0.20 [-0.81, 0.42] \\
dep. length $\sigma$ & 1.07$^{**}$ [0.42, 1.71] & 0.12 [-0.17, 0.41] & 0.23 [-0.18, 0.64] & -0.40 [-0.87, 0.07] \\
list of three & 2.69$^{***}$ [2.09, 3.29] & 1.14$^{***}$ [0.78, 1.50] & 1.60$^{***}$ [1.00, 2.20] & 0.55$^{*}$ [0.10, 1.00] \\
\addlinespace[2pt]
\multicolumn{5}{l}{\textbf{Verbs}} \\
$\it{align}$ & 3.14$^{***}$ [2.71, 3.57] & 1.51$^{***}$ [1.20, 1.82] & 1.92$^{***}$ [1.62, 2.22] & 2.71$^{***}$ [2.20, 3.22] \\
$\it{delve}$ & -5.52 [-10.73, -0.31] & -2.53 [-4.69, -0.37] & -2.43 [-4.94, 0.07] & -3.20 [-8.61, 2.21] \\
$\it{enhance}$ & 0.94 [-3.22, 5.11] & 0.96 [-0.99, 2.92] & 1.30 [-1.32, 3.92] & 1.47 [-0.24, 3.18] \\
$\it{exhibit}$ & 2.19$^{***}$ [1.44, 2.93] & 0.76$^{***}$ [0.52, 1.00] & 0.95$^{***}$ [0.64, 1.25] & 1.37$^{***}$ [0.75, 1.99] \\
$\it{foster}$ & -0.07 [-1.67, 1.53] & -0.09 [-0.74, 0.57] & -0.37 [-0.73, -0.02] & 0.24 [-0.16, 0.64] \\
$\it{garner}$ & -1.26 [-2.35, -0.16] & -0.38 [-0.77, 0.01] & -0.56 [-1.40, 0.29] & 0.13 [-0.04, 0.29] \\
$\it{showcase}$ & -1.96 [-3.92, 0.01] & -0.70 [-1.64, 0.23] & -1.23$^{*}$ [-2.14, -0.33] & -0.38 [-1.40, 0.64] \\
$\it{underscore}$ & 4.25 [-3.88, 12.38] & 1.46 [-1.22, 4.15] & 2.84 [-1.17, 6.84] & 4.53$^{***}$ [2.41, 6.65] \\
\bottomrule
\caption{Abstract-level interrupted-time-series results. Cells report the annualized post-intervention change in slope ($\beta_3$), standardized by the pre-intervention monthly standard deviation, with 95\% confidence intervals. Positive values indicate a shift toward a more positive trajectory; negative values indicate a shift toward a more negative trajectory. $^*q<.05$, $^{**}q<.01$, $^{***}q<.001$ after within-family Benjamini--Hochberg correction.}\label{tab:full-feature-its}\\

\end{longtable}
\endgroup

%% file: tables/excess.tex
\begingroup
\small
\setlength{\tabcolsep}{3.5pt}
\begin{longtable}{@{}lllll@{}}
\toprule
Feature & AI & q-bio & stat & medRxiv \\
\midrule
\endfirsthead
\toprule
Feature & AI & q-bio & stat & medRxiv \\
\midrule
\endhead
\multicolumn{5}{l}{\textbf{Adjectives}} \\
$\it{comprehensive}$ & 3.84$^{***}$ [2.08, 5.59] & 2.69$^{**}$ [1.13, 4.24] & 2.12$^{***}$ [1.09, 3.15] & 2.30$^{***}$ [1.24, 3.35] \\
$\it{crucial}$ & 3.39$^{***}$ [1.91, 4.87] & 2.45$^{**}$ [1.04, 3.85] & 1.95$^{***}$ [0.95, 2.96] & 3.84$^{***}$ [2.11, 5.58] \\
$\it{intricate}$ & 7.25$^{***}$ [3.41, 11.08] & 3.99$^{***}$ [2.31, 5.68] & 3.67$^{***}$ [1.80, 5.55] & 4.51$^{***}$ [2.76, 6.26] \\
$\it{pivotal}$ & 6.87$^{**}$ [2.39, 11.36] & 3.12$^{**}$ [1.19, 5.05] & 2.03$^{***}$ [1.16, 2.90] & 0.54$^{**}$ [0.15, 0.92] \\
$\it{potential}$ & 4.79$^{***}$ [3.34, 6.23] & 1.20$^{**}$ [0.46, 1.95] & 1.56$^{***}$ [0.71, 2.40] & 2.98$^{***}$ [1.77, 4.20] \\
\addlinespace[2pt]
\multicolumn{5}{l}{\textbf{Causal markers}} \\
$\it{accordingly}$ & -0.58$^{**}$ [-0.95, -0.22] & -0.43 [-0.80, -0.06] & -0.70$^{*}$ [-1.25, -0.15] & 0.19 [-0.04, 0.42] \\
$\it{consequently}$ & 0.90$^{***}$ [0.45, 1.34] & 0.30 [-0.32, 0.92] & 0.44$^{*}$ [0.04, 0.84] & 0.04 [-1.08, 1.15] \\
$\it{hence}$ & -1.76$^{***}$ [-2.68, -0.85] & -0.57 [-1.17, 0.03] & -1.54$^{***}$ [-2.14, -0.95] & -0.08 [-0.55, 0.39] \\
$\it{thereby}$ & 3.81$^{**}$ [1.23, 6.38] & 1.74$^{**}$ [0.80, 2.68] & 2.35$^{***}$ [1.20, 3.50] & 0.38 [-0.02, 0.77] \\
$\it{thus}$ & -1.53$^{***}$ [-2.30, -0.75] & -0.26 [-0.90, 0.37] & -0.97$^{***}$ [-1.47, -0.46] & -0.54 [-1.36, 0.29] \\
\addlinespace[2pt]
\multicolumn{5}{l}{\textbf{Contrast markers}} \\
$\it{alternatively}$ & -0.18 [-0.42, 0.06] & 0.18 [-0.33, 0.70] & -0.12 [-0.54, 0.30] & 0.95 [0.22, 1.67] \\
$\it{conversely}$ & 1.07$^{**}$ [0.39, 1.74] & 0.18 [-0.31, 0.66] & 0.69 [-0.10, 1.48] & 0.89 [-0.04, 1.83] \\
$\it{however}$ & 0.04 [-0.22, 0.30] & 0.58 [-0.15, 1.31] & -0.42 [-1.23, 0.39] & 0.44 [-0.21, 1.09] \\
$\it{nevertheless}$ & 0.10 [-0.48, 0.67] & -0.17 [-0.84, 0.50] & -0.02 [-0.66, 0.62] & -0.25 [-0.73, 0.23] \\
$\it{nonetheless}$ & 0.44 [0.00, 0.88] & -0.10 [-0.49, 0.30] & -0.10 [-0.59, 0.39] & -0.05 [-0.74, 0.64] \\
\addlinespace[2pt]
\multicolumn{5}{l}{\textbf{Emphasis markers}} \\
$\it{crucially}$ & 0.25 [-0.10, 0.60] & 0.20 [-0.20, 0.61] & 0.37 [-0.23, 0.96] & 0.33 [-0.07, 0.72] \\
$\it{importantly}$ & -0.16 [-0.45, 0.12] & 0.52$^{*}$ [0.08, 0.97] & 0.46 [-0.07, 1.00] & 0.27 [-0.11, 0.66] \\
$\it{notably}$ & 4.78$^{**}$ [2.21, 7.34] & 2.14$^{***}$ [1.36, 2.92] & 2.11$^{**}$ [0.99, 3.23] & 3.04$^{***}$ [1.50, 4.58] \\
$\it{particularly}$ & 4.41$^{**}$ [1.62, 7.21] & 1.84$^{***}$ [0.90, 2.78] & 2.63$^{**}$ [1.04, 4.21] & 1.84 [0.17, 3.51] \\
$\it{remarkably}$ & 0.64$^{**}$ [0.27, 1.01] & 0.29 [-0.25, 0.82] & 0.35 [-0.18, 0.88] & 0.39 [-0.45, 1.24] \\
\addlinespace[2pt]
\multicolumn{5}{l}{\textbf{Marker words}} \\
$\it{across}$ & 3.50$^{*}$ [0.77, 6.23] & 1.78$^{**}$ [0.52, 3.05] & 2.36$^{*}$ [0.59, 4.13] & 0.90 [-0.13, 1.93] \\
$\it{additionally}$ & 4.95$^{***}$ [2.63, 7.26] & 3.34$^{***}$ [2.00, 4.68] & 3.95$^{***}$ [2.52, 5.39] & 2.31$^{***}$ [1.23, 3.39] \\
$\it{finding}$ & 3.86$^{***}$ [2.24, 5.48] & 2.09$^{***}$ [1.19, 2.99] & 2.48$^{***}$ [1.46, 3.50] & 1.56$^{*}$ [0.12, 3.00] \\
$\it{insight}$ & 3.41$^{***}$ [1.86, 4.96] & 2.51$^{***}$ [1.29, 3.73] & 2.07$^{**}$ [0.64, 3.50] & 1.44$^{**}$ [0.57, 2.31] \\
$\it{interplay}$ & 0.69$^{**}$ [0.29, 1.10] & 0.53$^{*}$ [0.04, 1.03] & 0.51 [-0.22, 1.24] & 1.00$^{*}$ [0.21, 1.80] \\
$\it{meticulously}$ & 8.68$^{**}$ [2.54, 14.82] & 2.08$^{**}$ [0.71, 3.46] & 1.84$^{**}$ [0.63, 3.04] & 2.84$^{**}$ [1.01, 4.66] \\
$\it{notably}$ & 4.78$^{***}$ [2.21, 7.34] & 2.14$^{***}$ [1.36, 2.92] & 2.11$^{***}$ [0.99, 3.23] & 3.04$^{***}$ [1.50, 4.58] \\
$\it{particularly}$ & 4.41$^{**}$ [1.62, 7.21] & 1.84$^{***}$ [0.90, 2.78] & 2.63$^{**}$ [1.04, 4.21] & 1.84$^{*}$ [0.17, 3.51] \\
$\it{realm}$ & 4.19$^{**}$ [1.62, 6.76] & 1.69$^{**}$ [0.68, 2.71] & 1.82$^{**}$ [0.72, 2.93] & 0.81$^{*}$ [0.23, 1.39] \\
$\it{while}$ & 0.94$^{*}$ [0.17, 1.72] & 1.23$^{**}$ [0.52, 1.94] & 0.13 [-0.46, 0.71] & 0.74 [-0.12, 1.60] \\
$\it{within}$ & 4.32$^{**}$ [1.85, 6.79] & 1.46$^{***}$ [0.80, 2.13] & 3.23$^{***}$ [2.02, 4.43] & -0.26 [-0.87, 0.35] \\
\addlinespace[2pt]
\multicolumn{5}{l}{\textbf{Other}} \\
certainty terms & 3.43$^{***}$ [1.69, 5.18] & 2.46$^{***}$ [1.34, 3.58] & 1.50$^{**}$ [0.50, 2.51] & 1.98$^{**}$ [0.84, 3.12] \\
hedge terms & 0.09 [-0.28, 0.46] & -0.94$^{*}$ [-1.78, -0.10] & -0.89$^{**}$ [-1.50, -0.28] & 0.91$^{*}$ [0.11, 1.71] \\
\addlinespace[2pt]
\multicolumn{5}{l}{\textbf{Phrases}} \\
$\it{comprehensive\ chapter}$ & -- & -- & -- & -- \\
$\it{deep\ dive}$ & 0.02 [-0.88, 0.92] & -0.09 [-1.01, 0.83] & -0.14 [-0.49, 0.21] & 0.53 [-0.17, 1.23] \\
$\it{essential\ insight}$ & -- & -- & -- & -- \\
$\it{intricate\ interplay}$ & 0.85 [0.18, 1.52] & 0.62 [-0.12, 1.36] & 3.32 [0.72, 5.92] & 9.61$^{***}$ [5.13, 14.09] \\
$\it{intricate\ web}$ & -- & 0.01 [-0.08, 0.10] & -- & -0.22 [-0.82, 0.38] \\
$\it{meticulously\ delve}$ & -- & -- & -- & -- \\
\addlinespace[2pt]
\multicolumn{5}{l}{\textbf{Punctuation}} \\
colon & 0.95$^{**}$ [0.39, 1.51] & 0.25 [-0.18, 0.69] & 0.01 [-0.53, 0.54] & -0.76 [-1.66, 0.13] \\
em dash & -0.21 [-0.59, 0.18] & -0.34 [-0.91, 0.24] & -0.04 [-0.69, 0.61] & 1.06 [-0.02, 2.13] \\
semicolon & -0.65$^{*}$ [-1.14, -0.17] & -0.47 [-1.06, 0.13] & -1.02 [-1.99, -0.05] & -0.96 [-1.83, -0.09] \\
\addlinespace[2pt]
\multicolumn{5}{l}{\textbf{Readability}} \\
ARI & 1.75$^{**}$ [0.72, 2.78] & 2.28$^{**}$ [0.85, 3.70] & 1.41$^{***}$ [0.78, 2.04] & -0.80 [-1.80, 0.21] \\
syllables/word & 3.80$^{**}$ [1.58, 6.02] & 3.47$^{***}$ [1.68, 5.27] & 3.75$^{***}$ [2.25, 5.24] & -0.22 [-1.41, 0.98] \\
Dale-Chall & 1.84$^{**}$ [0.73, 2.96] & 3.08$^{***}$ [1.43, 4.73] & 2.96$^{***}$ [1.71, 4.21] & -0.40 [-1.55, 0.76] \\
FKGL & 2.51$^{**}$ [1.07, 3.96] & 2.33$^{**}$ [0.90, 3.76] & 1.97$^{***}$ [1.01, 2.92] & -0.32 [-1.09, 0.46] \\
FRE & -2.67$^{**}$ [-4.24, -1.09] & -3.14$^{***}$ [-4.91, -1.36] & -2.70$^{***}$ [-3.87, -1.54] & 0.05 [-0.86, 0.96] \\
Gunning Fog & 3.27$^{**}$ [1.48, 5.05] & 2.55$^{***}$ [1.18, 3.91] & 2.37$^{***}$ [1.11, 3.62] & -0.20 [-0.96, 0.56] \\
SMOG & 2.55$^{**}$ [1.17, 3.94] & 2.13$^{***}$ [0.95, 3.31] & 2.08$^{***}$ [0.94, 3.21] & -0.48 [-1.16, 0.21] \\
\addlinespace[2pt]
\multicolumn{5}{l}{\textbf{Sequential markers}} \\
$\it{additionally}$ & 4.95$^{***}$ [2.63, 7.26] & 3.34$^{***}$ [2.00, 4.68] & 3.95$^{***}$ [2.52, 5.39] & 2.31$^{***}$ [1.23, 3.39] \\
$\it{further}$ & 1.30$^{***}$ [0.79, 1.80] & 0.13 [-0.37, 0.62] & 0.09 [-0.44, 0.63] & 0.70$^{*}$ [0.22, 1.19] \\
$\it{furthermore}$ & 1.11$^{***}$ [0.58, 1.65] & 0.64$^{*}$ [0.18, 1.10] & 1.06$^{**}$ [0.42, 1.70] & -0.03 [-0.61, 0.55] \\
$\it{moreover}$ & -0.31 [-0.93, 0.32] & -0.44 [-1.23, 0.34] & -0.51 [-1.23, 0.21] & -0.18 [-1.22, 0.86] \\
$\it{subsequently}$ & 3.10$^{***}$ [1.57, 4.63] & 0.91$^{***}$ [0.41, 1.41] & 1.14$^{***}$ [0.55, 1.74] & 0.14 [-0.48, 0.76] \\
\addlinespace[2pt]
\multicolumn{5}{l}{\textbf{Summary markers}} \\
$\it{collectively}$ & 0.78$^{***}$ [0.51, 1.04] & 0.40 [0.01, 0.79] & 0.21 [-0.32, 0.74] & 0.56 [-0.22, 1.34] \\
$\it{in\ summary}$ & 0.79$^{***}$ [0.46, 1.11] & 0.28 [-0.20, 0.77] & 0.50$^{*}$ [0.08, 0.91] & -0.46 [-1.52, 0.60] \\
$\it{overall}$ & 0.49$^{**}$ [0.17, 0.81] & -0.10 [-0.61, 0.40] & -0.24 [-0.77, 0.30] & -0.62 [-1.31, 0.06] \\
$\it{taken\ together}$ & -0.12 [-0.42, 0.18] & 0.20 [-0.28, 0.69] & -0.38 [-0.76, 0.00] & -1.01$^{*}$ [-1.68, -0.34] \\
$\it{ultimately}$ & 1.75$^{***}$ [1.15, 2.36] & 0.42 [0.06, 0.79] & 0.67$^{*}$ [0.11, 1.23] & 0.98 [0.12, 1.84] \\
\addlinespace[2pt]
\multicolumn{5}{l}{\textbf{Syntax}} \\
words/sent. & -0.99$^{*}$ [-1.69, -0.29] & -1.03$^{**}$ [-1.58, -0.48] & -0.67 [-1.20, -0.14] & -1.91$^{***}$ [-2.81, -1.00] \\
max tree depth & -0.63 [-1.51, 0.25] & -0.85$^{*}$ [-1.46, -0.24] & -0.10 [-0.98, 0.77] & -1.13$^{*}$ [-2.02, -0.24] \\
tree depth $\sigma$ & -0.61 [-1.44, 0.22] & -0.40 [-1.01, 0.21] & 0.03 [-0.88, 0.95] & -0.57 [-1.48, 0.35] \\
coord. count & 0.13 [-0.12, 0.37] & -0.23 [-0.52, 0.06] & -0.40 [-0.74, -0.05] & -1.25$^{**}$ [-2.00, -0.50] \\
coord./sent. $\sigma$ & -0.10 [-0.60, 0.40] & -0.47 [-1.03, 0.09] & -0.50 [-0.94, -0.06] & -2.10$^{***}$ [-3.07, -1.13] \\
dep. entropy & -0.71$^{**}$ [-1.14, -0.28] & -0.72$^{*}$ [-1.28, -0.16] & -1.37$^{**}$ [-2.16, -0.58] & -1.04$^{*}$ [-1.96, -0.11] \\
dep. length & -0.20 [-0.57, 0.17] & -0.34 [-0.73, 0.05] & -0.14 [-0.59, 0.31] & -1.82$^{*}$ [-3.21, -0.44] \\
dep. length $\sigma$ & -0.36 [-0.84, 0.13] & -0.39 [-0.81, 0.04] & -0.29 [-0.98, 0.41] & -1.07 [-2.12, -0.02] \\
list of three & 1.19$^{***}$ [0.65, 1.74] & 0.24 [-0.43, 0.92] & 0.29 [-0.24, 0.83] & -0.11 [-0.99, 0.78] \\
\addlinespace[2pt]
\multicolumn{5}{l}{\textbf{Verbs}} \\
$\it{align}$ & 3.17$^{**}$ [1.07, 5.27] & 1.93$^{***}$ [0.97, 2.88] & 2.29$^{***}$ [0.97, 3.62] & 2.88$^{**}$ [1.06, 4.71] \\
$\it{delve}$ & 12.90$^{**}$ [5.47, 20.33] & 5.51$^{***}$ [3.00, 8.01] & 6.85$^{***}$ [3.70, 10.00] & 12.18$^{**}$ [5.12, 19.24] \\
$\it{enhance}$ & 8.41$^{**}$ [3.09, 13.72] & 5.03$^{***}$ [2.25, 7.80] & 5.83$^{***}$ [2.51, 9.15] & 3.96$^{**}$ [1.47, 6.45] \\
$\it{exhibit}$ & 4.89$^{***}$ [2.95, 6.82] & 1.10$^{***}$ [0.45, 1.74] & 2.01$^{***}$ [1.26, 2.77] & 2.55$^{***}$ [1.20, 3.90] \\
$\it{foster}$ & 2.91$^{**}$ [1.26, 4.55] & 1.25$^{***}$ [0.56, 1.95] & 0.81$^{*}$ [0.19, 1.43] & 0.24 [-0.18, 0.66] \\
$\it{garner}$ & 3.84$^{***}$ [2.64, 5.04] & 1.05$^{**}$ [0.30, 1.81] & 2.47$^{***}$ [1.52, 3.42] & 0.68$^{***}$ [0.32, 1.03] \\
$\it{showcase}$ & 4.09$^{**}$ [1.61, 6.57] & 1.98$^{***}$ [0.89, 3.08] & 2.49$^{***}$ [1.54, 3.44] & 1.97$^{**}$ [0.72, 3.23] \\
$\it{underscore}$ & 19.16$^{**}$ [7.27, 31.05] & 7.05$^{***}$ [3.26, 10.84] & 11.27$^{***}$ [6.67, 15.86] & 7.67$^{**}$ [3.17, 12.17] \\
\bottomrule
\caption{Abstract-level counterfactual-excess results for 2023--2024. Cells report the paper-count-weighted mean difference between observed monthly values and projections from the pre-intervention trend, standardized by the pre-intervention monthly standard deviation, with 95\% confidence intervals. $^*q<.05$, $^{**}q<.01$, $^{***}q<.001$ after within-family Benjamini--Hochberg correction.}\label{tab:full-feature-excess-2023-2024}\\

\end{longtable}
\endgroup